\documentclass[%
 reprint,
superscriptaddress,
 amsmath,amssymb,
 aps,
]{revtex4-2}

\usepackage{graphicx}
\usepackage{dcolumn}
\usepackage{booktabs}
\usepackage{bm}
\usepackage{fontawesome5}
\usepackage{amssymb}
\usepackage{hyperref}
\usepackage[english]{babel}  

\begin{document}

\preprint{APS/123-QED}
\title{Minimal Embodiment Enables Efficient Learning of Number Concepts in Robot}

\author{Zhegong Shangguan}
\thanks{Corresponding author: zhegong.shangguan@manchester.ac.uk}
\affiliation{%
Cognitive Robotics Lab, Department of Computer Science, University of Manchester
}%

\author{Alessandro Di Nuovo}
\affiliation{%
Smart Interactive Technologies (SIT) Research Laboratory, Department of Computing, Sheffield Hallam University
}%

\author{Angelo Cangelosi}
\affiliation{%
Cognitive Robotics Lab, Department of Computer Science, University of Manchester
}%

\begin{abstract}
Robots are increasingly entering human-interactive scenarios that require understanding of quantity. How intelligent systems acquire abstract numerical concepts from sensorimotor experience remains a fundamental challenge in cognitive science and artificial intelligence. Here we investigate embodied numerical learning using a neural network model trained to perform sequential counting through naturalistic robotic interaction with a Franka Panda manipulator. We demonstrate that embodied models achieve 96.8\% counting accuracy with only 10\% of training data, compared to 60.6\% for vision-only baselines. This advantage persists when visual-motor correspondences are randomized, indicating that embodiment functions as a structural prior that regularizes learning rather than as an information source. The model spontaneously develops biologically plausible representations: number-selective units with logarithmic tuning, mental number line organization, Weber-law scaling, and rotational dynamics encoding numerical magnitude ($r = 0.97$, slope $= 30.6°$/count). The learning trajectory parallels children's developmental progression from subset-knowers to cardinal-principle knowers. These findings demonstrate that minimal embodiment can ground abstract concepts, improve data efficiency, and yield interpretable representations aligned with biological cognition, which may contribute to embodied mathematics tutoring and safety-critical industrial applications.
\end{abstract}

\maketitle


\section{Introduction}\label{sec1}


How embodiment shapes abstract concept learning remains a fundamental question in cognitive science \cite{borghi2017challenge} and artificial intelligence \cite{pmlr-v267-li25p}. Numerical cognition, the ability to perceive, represent, and manipulate quantities, provides a tractable domain for investigating this question \cite{nieder2025calculating}. Human infants develop counting abilities through embodied interactions with their environment~\cite{piaget1965child,wynn1998psychological}, suggesting that sensorimotor grounding may play a critical role in abstract reasoning.

Converging evidence supports the embodied nature of numerical cognition\cite{wynn1992addition}. Children spontaneously use fingers when learning to count, and these gestures improve accuracy and support coordination of number words with objects~\cite{alibali1999function,roesch2024finger}. Neuroimaging reveals that the intraparietal sulcus shows overlapping activation for number representation and spatial attention, with arm kinematics systematically biasing numerical judgments~\cite{dehaene2005three,hubbard2005interactions,fischer2008embodied}. These findings raise a fundamental computational question: \textbf{does sensorimotor coupling provide a structured prior that shapes learning dynamics, or does it simply offer additional information channels?}

\begin{figure*}
    \centering
    \includegraphics[width=\linewidth]{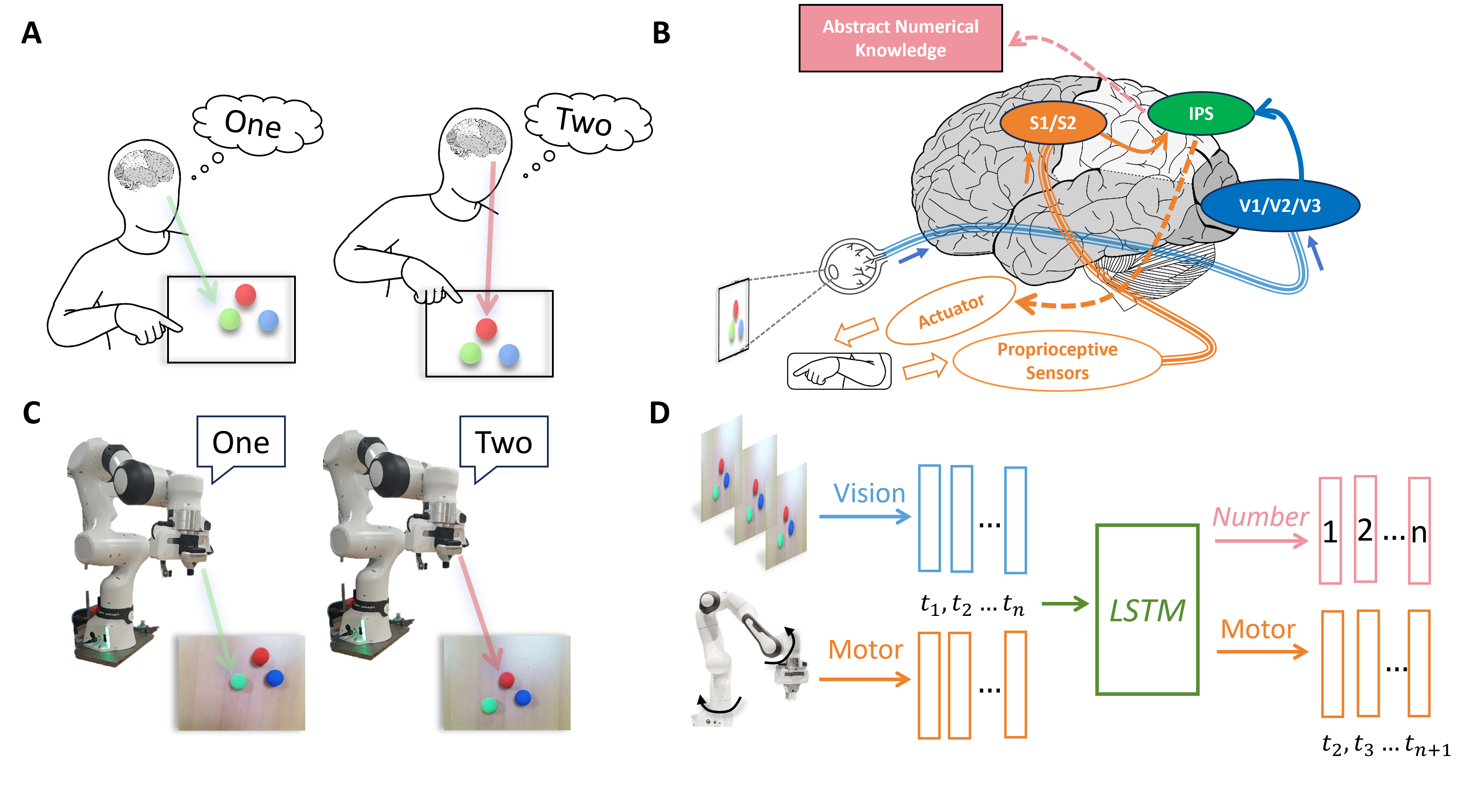}
    \caption{\textbf{Embodied numerical cognition: from human to robot.}
\textbf{(A)} Humans use pointing gestures to coordinate counting with objects.
\textbf{(B)} Neural basis: visual (V1/V2/V3) and proprioceptive (S1/S2) signals converge in the intraparietal sulcus (IPS), supporting abstract numerical knowledge.
\textbf{(C)} Robotic implementation: a Franka Panda manipulator performs sequential counting with a wrist-mounted camera.
\textbf{(D)} Model architecture: visual and motor inputs are integrated by an LSTM, which outputs number classification and next-step motor prediction.}
\label{fig:overview}
\end{figure*}

The acquisition of counting follows a structured developmental path. Children progress from subset-knowers, who grasp only small numerals (typically 1--3), to cardinal-principle knowers (CP--knowers) who generalize counting across their entire numerical range~\cite{gelman1986child,
le2007one}. This pattern is consistent across languages~\cite{le2016numerical}, suggesting 
universal computational constraints on numerical learning.

Computational modeling has revealed different aspects of numerical cognition. Stoianov and Zorzi trained a hierarchical generative model on visual images and found that number-selective neurons emerged in the deepest layer, with monotonic encoding consistent with single-cell recordings in monkey parietal cortex~\cite{stoianov2012emergence}. Kim et al. found similar tuning in untrained convolutional neural networks (CNNs), suggesting that innate number sense may originate from the statistical properties of feedforward visual processing~\cite{kim2021visual}. These findings reveal that basic numerosity representations can arise from visual statistics in single images. A separate line of work has examined counting as a sequential process. Rodriguez and Wiles demonstrated that recurrent neural networks (RNNs) can implement symbol-sensitive counting~\cite{rodriguez1997recurrent}, and Fang et al. showed that long short-term memory (LSTM) networks exhibit learning trajectories resembling children's subset-knower stages~\cite{fang2018can}.

Cognitive developmental robotics bridges these approaches by grounding counting in embodied interaction~\cite{Cangelosi2022}. Pecyna et al. demonstrated that pointing gestures facilitate counting acquisition in the iCub humanoid robot~\cite{Pecyna2020}. Di Nuovo and McClelland showed that finger counting boosts number learning: proprioceptive information from robot hands improved digit recognition and supported quicker formation of a uniform number line~\cite{dinuovo2019developing}. Notably, they found embodied models perform better than baseline even when motor signals were not aligned with visual input, suggesting that embodiment may contribute something beyond task-relevant information. \textbf{But do these effects depend on humanoid morphology, or do they reflect general principles of embodiment?}

Here, we address these questions using an embodied neural network combining CNNs for visual processing with LSTMs for sequential counting. We collected egocentric visual and motor data from a Franka Panda robotic manipulator~\cite{ELSNER2023101532} performing naturalistic counting tasks, and trained the model using supervised learning (Figure~\ref{fig:overview}). Our approach has three key elements. First, we employ a non-humanoid platform to test whether embodiment effects generalize beyond human-like morphology. Second, we compare intact versus shuffled visual-motor correspondences to test whether veridical sensorimotor coupling is necessary. Third, we apply analytical methods from systems neuroscience, including jPCA~\cite{churchland2012neural}, to characterize the internal dynamics underlying counting.

Our experiments reveal that embodied models achieve 96.8\% accuracy with only 10\% of training data, compared to 60.6\% for vision-only baselines. This advantage persists even when visual-motor correspondences are randomized, demonstrating that embodiment functions as a structural prior rather than an information source. We term this ``minimalist embodiment.'' Analysis of neural population activity reveals that the model spontaneously develops biologically plausible representations: number-selective units with logarithmic tuning and mental number line organization in representational space. Strikingly, jPCA analysis uncovers rotational dynamics during counting, with the phase of rotation encoding numerical magnitude ($r = 0.97$). This suggests that the network implements counting as a single rotational cycle through neural state space. By combining embodied robotics with analytical methods from cognitive neuroscience, we show that artificial neural networks can develop interpretable, human-aligned representations for abstract concepts. These findings bridge machine learning, developmental robotics, and systems neuroscience, pointing toward AI systems that learn more efficiently through embodied interaction. These results may also offer insights into the role of embodiment in children's early numerical learning.

\section{Results}\label{sec2}

\begin{figure*}[htbp]
\centering
\includegraphics[width=\textwidth]{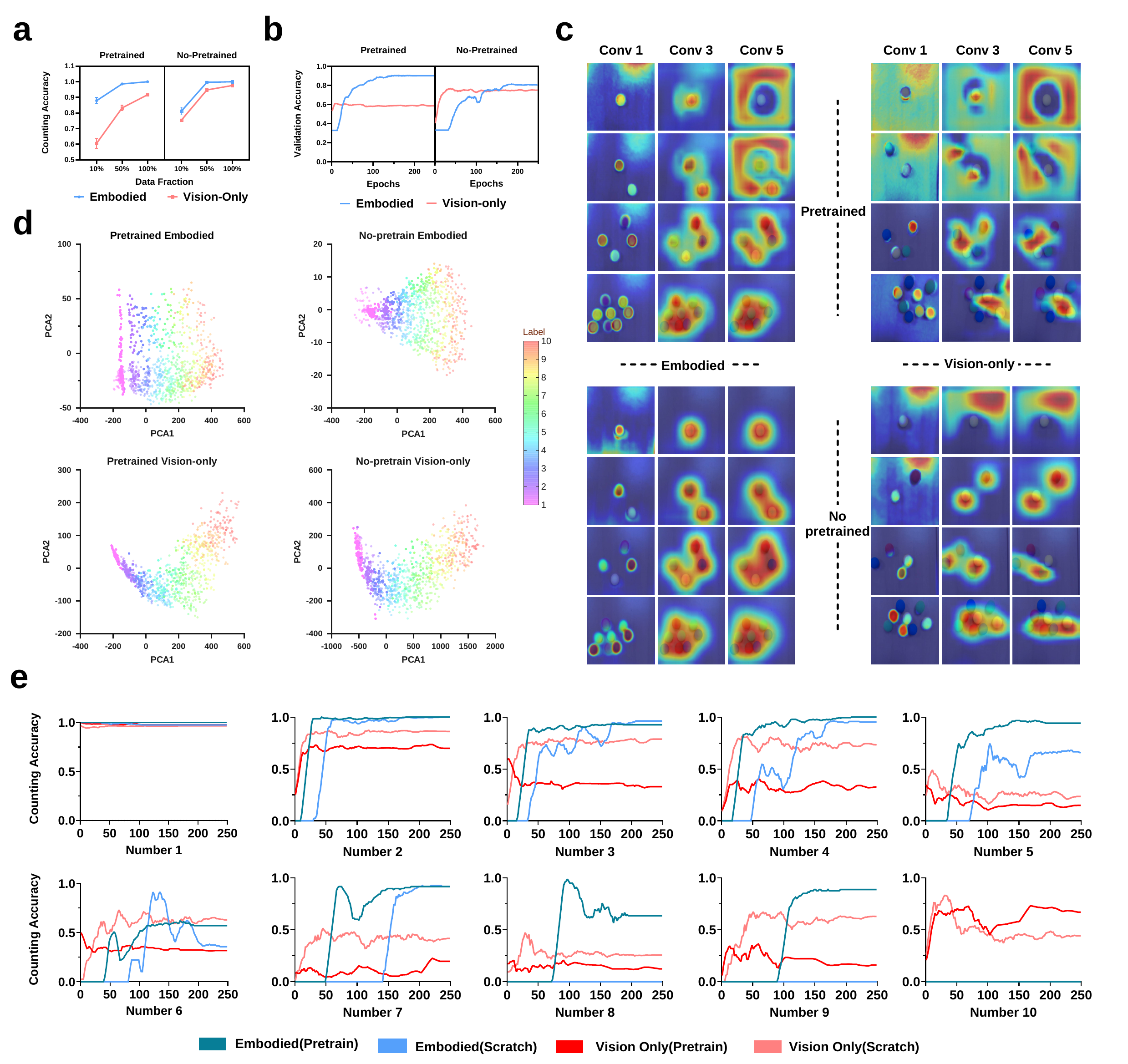}
\caption{\textbf{Embodied learning yields superior data efficiency, distinctive learning dynamics, and structured numerical representations.}
\textbf{a,} Validation accuracy across training data fractions (10\%, 50\%, 100\%) for embodied (blue) and vision-only (red) models, with and without ImageNet pretraining. Embodied models consistently outperform vision-only baselines across all conditions. Notably, pretraining enhances embodied model performance while impairing vision-only models, a negative transfer effect most pronounced in low-data regimes. Error bars indicate standard deviation across 3 random seeds.
\textbf{b,} Learning trajectories over training epochs (10\% data condition). The vision-only model (red) exhibits rapid early improvement followed by plateau, whereas embodied models (blue) show an initial latent period (20--50 epochs) before sharp performance gains that ultimately surpass vision-only baselines. Shaded regions indicate standard deviation.
\textbf{c,} Grad-CAM visualizations of visual attention across convolutional layers (Conv1, Conv3, Conv5) for numerosities 1, 2, 4, and 8. Embodied models maintain focused attention on target objects across all conditions, while vision-only models, particularly with pretraining, allocate attention broadly to background regions, especially for small numerosities.
\textbf{d,} PCA visualization of visual encoder (AlexNet) representations, with colors indicating numerosity (1--10, purple to yellow). Pretrained embodied models develop linearly organized representations resembling a mental number line, while vision-only models exhibit distorted U-shaped geometry.
\textbf{e,} Per-number learning trajectories across training epochs. Embodied models (blue shades) acquire numbers sequentially (1\(\rightarrow\)2\(\rightarrow\)3\(\rightarrow\)...), mirroring children's developmental progression from subset-knowers to cardinal-principle knowers. Vision-only models (red shades) show approximate discrimination across all numbers from early training, resembling the Approximate Number System (ANS). All panels show results from models trained on 10\% of the full dataset. Vision-only results are from the single-image baseline (V1 and V3); the sequential-image baseline (V2 and V4) yielded nearly identical performance (see Table~\ref{tab:val_best}).}
\label{fig:performance_comparison}
\end{figure*}

\begin{table*}[ht]
\centering
\caption{Validation performance at best checkpoint. Metrics shown as mean (std) over 3 seeds. Accuracy is final\_val\_count\_accuracy (\%), MSE is joint MSE (\(\times 10^2\)). Bold indicates best per data fraction and metric.}
\label{tab:val_best}
\resizebox{\textwidth}{!}{%
\begin{tabular}{lccc ccccc ccccc}
\toprule
\textbf{Model} 
& \multicolumn{3}{c}{\textbf{Conditions}} 
& \multicolumn{2}{c}{\textbf{10\% Data}} 
& \multicolumn{2}{c}{\textbf{50\% Data}} 
& \multicolumn{2}{c}{\textbf{100\% Data}} \\
\cmidrule(r){2-4} \cmidrule(r){5-6} \cmidrule(r){7-8} \cmidrule(r){9-10}
& \textbf{Pretrain} & \textbf{Joint Shuffle} & \textbf{Curriculum}
& Acc (\%) $\uparrow$ & MSE $\downarrow$
& Acc (\%) $\uparrow$ & MSE $\downarrow$
& Acc (\%) $\uparrow$ & MSE $\downarrow$ \\
\midrule

E1 & $\times$ & $\times$ & Easy\(\rightarrow\)Hard 
& 85.7(2.7) & 1.73(0.04)
& 97.2(0.2) & 1.54(0.04)
& 95.8(0.7) & 1.77(0.09) \\

E2 & $\times$ & $\times$ & Hard\(\rightarrow\)Easy 
& 84.0(3.3) & 1.74(0.04)
& 96.3(0.2) & 1.67(0.08)
& 94.2(0.4) & 1.87(0.21) \\

E3 & $\times$ & $\times$ & Random 
& 90.1(2.7) & 1.70(0.03)
& 99.9(0.1) & 1.68(0.08)
& 100.0(0.0) & 1.05(0.15) \\

\cmidrule(r){1-10}

E4 & $\times$ & \checkmark & Easy\(\rightarrow\)Hard 
& 85.5(3.3) & 2.05(0.05)
& 97.2(0.4) & 1.97(0.07)
& 97.2(1.7) & 2.19(0.07) \\

E5 & $\times$ & \checkmark & Hard\(\rightarrow\)Easy 
& 84.0(3.7) & 2.10(0.03)
& 85.2(18.1) & 2.11(0.14)
& 82.1(27.1) & 2.31(0.10) \\

E6 & $\times$ & \checkmark & Random 
& 90.2(2.5) & 2.08(0.02)
& 99.9(0.02) & 2.21(0.08)
& 100.0(0.1) & 1.71(0.09) \\

\cmidrule(r){1-10}

V1 & $\times$ & $\times$ & Random 
& 75.4(0.01) & --
& 94.7(0.01) & --
& 97.6(0.01) & -- \\

V2 & $\times$ & $\times$ & Random 
& 77.9(0.01) & --
& 94.8(0.01) & --
& 96.3(0.01) & -- \\

\cmidrule(r){1-10}

E7 & \checkmark & $\times$ & Easy\(\rightarrow\)Hard 
& 94.7(0.8) & 1.42(0.15)
& 99.1(0.1) & 1.39(0.28)
& 99.1(0.9) & 1.33(0.44) \\

E8 & \checkmark & $\times$ & Hard\(\rightarrow\)Easy 
& 94.1(0.1) & 1.46(0.14)
& 98.6(0.1) & 1.08(0.07)
& 96.3(6.1) & 1.03(0.15) \\

E9 & \checkmark & $\times$ & Random 
& 96.8(0.9) & 1.46(0.17)
& 99.9(0.0) & 0.69(0.09)
& 100.0(0.0) & 0.74(0.13) \\

\cmidrule(r){1-10}

E10 & \checkmark & \checkmark & Easy\(\rightarrow\)Hard 
& 94.3(0.7) & 1.95(0.14)
& 99.2(0.2) & 1.87(0.29)
& 99.1(1.1) & 1.69(0.44) \\

E11 & \checkmark & \checkmark & Hard\(\rightarrow\)Easy 
& 94.0(0.3) & 1.97(0.20)
& 98.5(0.3) & 1.59(0.10)
& 85.9(24.1) & 1.90(0.28) \\

E12 & \checkmark & \checkmark & Random 
& 96.6(0.7) & 2.02(0.06)
& 99.8(0.2) & 1.35(0.15)
& 100.0(0.0) & 1.39(0.08) \\

\cmidrule(r){1-10}

V3 & \checkmark & $\times$ & Random 
& 60.6(0.03) & --
& 83.3(0.01) & --
& 91.6(0.01) & -- \\

V4 & \checkmark & $\times$ & Random 
& 61.0(0.01) & --
& 81.8(0.00) & --
& 91.9(0.02) & -- \\

\bottomrule
\end{tabular}
}
\end{table*}

\subsection{Embodied learning enhances counting performance and data efficiency}
\label{subsection A}
\begin{figure*}
    \centering
    \includegraphics[width=.5\linewidth]{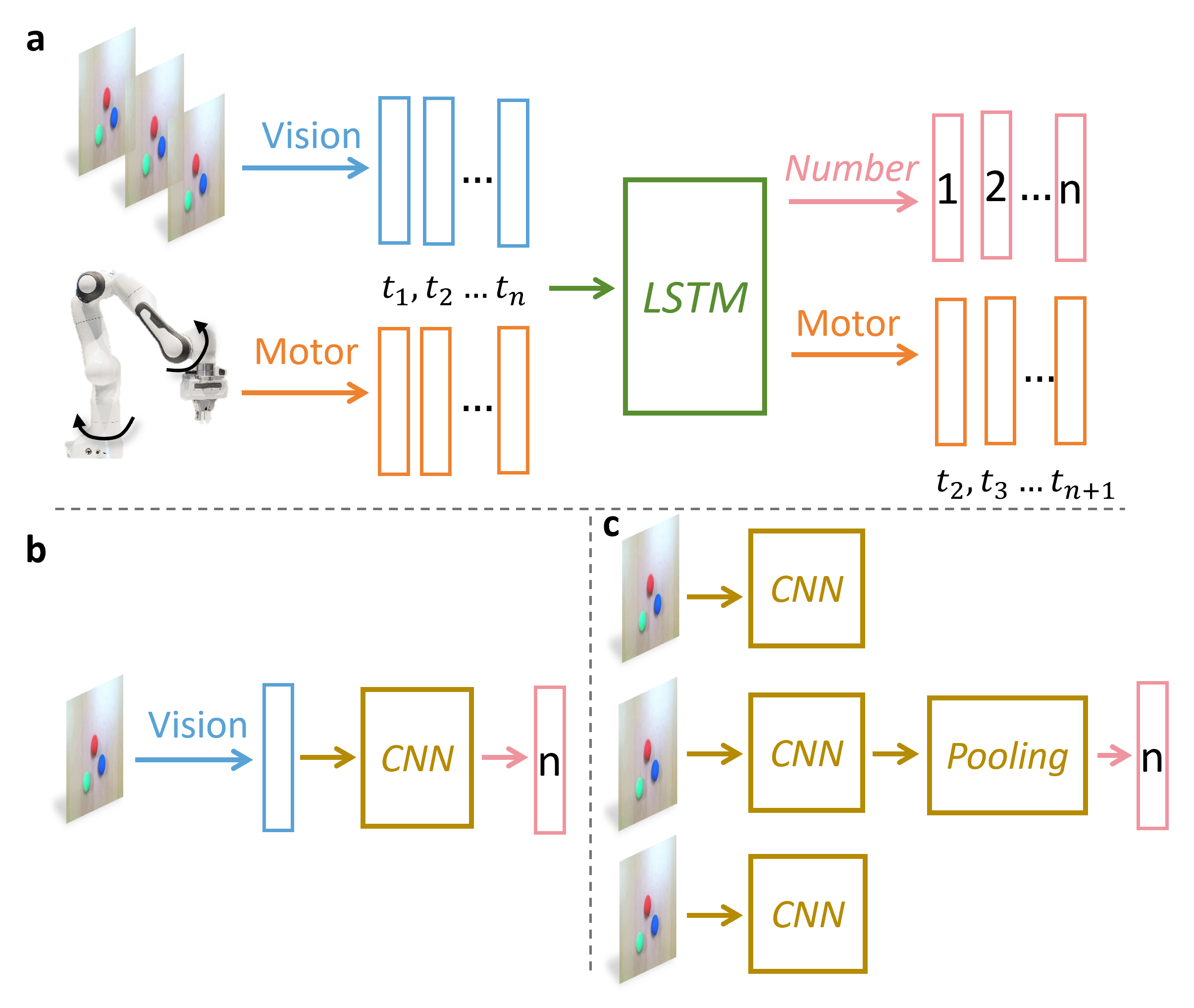}
    \caption{Overview of three modeling paradigms: (a) an embodied model that integrates visual observations and motor signals over time using an LSTM to predict sequential states and actions; (b) single-image classification using a CNN; and (c) sequential-image classification, where multiple frames are processed by CNNs and aggregated (e.g., via pooling) to produce a final prediction.}
    \label{fig: apeedix2}
\end{figure*}

To investigate the computational principles underlying embodied numerical learning, we trained an integrated neural network model combining convolutional neural networks (CNNs) for visual processing with long short-term memory (LSTM) networks for sequential counting (Figure~\ref{fig:overview}C-D). The model learned from a robotic counting task in which a Franka Panda manipulator sequentially visited objects arranged in its workspace while a wrist-mounted camera captured egocentric images at each step (Supply Material Figure~\ref{fig: apeedix1}). We systematically compared this embodied approach against two vision-only baselines that processed identical visual inputs without motor information: a single-image classifier that predicted count from frames, and a sequential-image model that aggregated features across sequential counting frames via mean pooling (Supply Material Figure~\ref{fig: apeedix2}). We also examined the effects of visual encoder pretraining (ImageNet-initialized versus randomly initialized) across different data regimes (10\%, 50\%, and 100\% of training data). We compared embodied models against vision-only baselines across pretraining conditions (see Table~\ref{tab:val_best} for full model specifications). Because single-image and sequential-image baselines yielded nearly identical performance, we focus subsequent analyses on embodied versus single-image comparisons.

Embodied learning demonstrated substantial advantages in counting accuracy across all data conditions (Figure~\ref{fig:performance_comparison}a; Table~\ref{tab:val_best}). In the most data-limited condition (10\% of training data), the embodied model with pretrained visual features (E9) achieved 96.8\% ($\pm$0.9\%) accuracy, while the embodied model without pretraining (E3) reached 90.1\% ($\pm$2.7\%). The vision-only baseline performed considerably worse under the same conditions: V1 achieved 75.4\% ($\pm$0.01\%) without pretraining, while V3 with pretraining showed further degraded performance at 60.6\% ($\pm$0.03\%). This pattern reveals an interaction between embodiment and visual pretraining: ImageNet initialization enhanced embodied model performance while substantially impairing vision-only models, corresponding to a negative transfer of approximately 15 percentage points. This asymmetry suggests that powerful visual representations trained for object recognition may introduce biases misaligned with numerical discrimination, particularly for small quantities where target objects occupy minimal visual space. Motor grounding appears to counteract these biases, enabling the embodied model to leverage pretrained features effectively. As training data increased, all models improved, yet the embodied advantage persisted: at 50\% and 100\% data, embodied models consistently outperformed vision-only baselines regardless of pretraining condition, with the gap most pronounced when pretrained encoders were used.

Analysis of learning dynamics over training epochs revealed qualitatively distinct developmental trajectories between architectures (Figure~\ref{fig:performance_comparison}b). The vision-only model exhibited rapid initial improvement, reaching approximately 60\% accuracy within the first 20 epochs, but subsequently plateaued with limited further gains despite continued training. In contrast, embodied models followed a markedly different trajectory characterized by an extended latent period: performance remained near chance for approximately 20--50 epochs before undergoing a sharp transition to rapid improvement, ultimately surpassing vision-only performance and approaching ceiling accuracy. This two-phase learning profile, an initial plateau followed by sudden competence, suggests that embodied counting requires an initial period to establish stable visuomotor alignment before numerical learning can proceed. Notably, the sequential-image baseline, which aggregates visual features across multiple frames, showed learning dynamics nearly identical to the single-image model, indicating that temporal visual information alone, without motor grounding, provides no additional benefit for this task. This pattern resonates with developmental observations that children's counting competence emerges gradually after extended periods of seemingly unproductive practice \cite{wynn1992children, le2007one}, during which foundational sensorimotor-symbolic mappings are being established. Learning trajectories for all models are shown in Supplementary Figure~\ref{fig: apeedix3}.

Gradient-weighted class activation mapping (Grad-CAM) analysis revealed systematic differences in visual attention that illuminate the mechanistic basis of these performance gaps (Figure~\ref{fig:performance_comparison}c). We visualized attention patterns across convolutional layers (Conv1, Conv3, Conv5) for both model types under pretrained and non-pretrained conditions, examining responses to small (1, 2 balls) and larger (4, 8 balls) numerosities. Embodied models, regardless of pretraining status, exhibited focused attention on target objects from early processing stages: Conv1 and Conv3 activations concentrated on ball locations, while Conv5 developed more regionalized, spatially coherent attention patterns. When pretrained features were used, embodied models showed slightly broader attention at Conv5 for small numerosities ($<$4), with some activation spreading to background regions, yet maintained predominant focus on task-relevant objects.

Vision-only models displayed qualitatively different attention patterns that varied systematically with numerosity. Without pretraining, these models could attend to scenes containing two or more balls but struggled with single-ball displays, where attention diffused broadly across the visual field. With pretraining, this pattern became more pronounced: attention was heavily allocated to background regions, particularly for small numerosities, with activations distributed diffusely across Conv1, Conv3, and Conv5 layers. For larger numerosities, vision-only models faced a different challenge, namely difficulty encompassing the full spatial extent of multi-object displays. This pattern suggests that ImageNet pretraining, optimized for recognizing coherent objects and scenes, may bias representations toward background context and global scene statistics rather than discrete object enumeration. The embodied model's motor signals appear to function as an attentional anchor, constraining visual processing toward task-relevant regions even when pretrained features introduce competing biases. Full visualizations and data are available in the online supplementary materials\footnote{OSF repository: \url{https://osf.io/jk4u8/overview?view_only=95fcde69554045788995b8ab2fdabc0d}}.

Principal component analysis (PCA) of visual encoder outputs revealed clear differences in representational geometry that further distinguish embodied from vision-only learning (Figure~\ref{fig:performance_comparison}d, upper row). The pretrained embodied model developed representations in which numerical magnitudes were arranged along a clear linear trajectory from left to right in PCA space, with small numbers forming tight clusters and larger numbers extending progressively along this axis. This organization closely resembles the ``mental number line'' posited in cognitive theories of human numerical representation \cite{dehaene2005three, de2014representations}, a spatial mapping in which numerical magnitude corresponds to position along an ordered continuum. The non-pretrained embodied model exhibited a radial organization that, while less strictly linear, nonetheless preserved ordinal structure with numbers arranged in a coherent sequence.

In contrast, both vision-only models, regardless of pretraining, produced U-shaped representational geometries in which numbers were arranged along a curved manifold rather than a linear axis. Although numerical order was partially preserved along this U-shaped trajectory (Figure~\ref{fig:performance_comparison}d, lower row), such geometry is suboptimal for magnitude discrimination: items at opposite ends of the numerical range (e.g., 1 and 10) may occupy nearby positions in representational space, while adjacent numbers near the curve's inflection point become maximally separated. This geometric distortion likely contributes to the poorer generalization observed in vision-only models, as linear representations support straightforward magnitude comparison through simple distance computations, whereas curved representations require more complex, potentially less robust decision boundaries.

Per-number analysis of learning trajectories revealed developmental patterns that parallel children's acquisition of counting competence (Figure~\ref{fig:performance_comparison}e). Across all model types, small numbers (1--4) were acquired more readily than large numbers, consistent with the well-documented subitizing range in which humans and animals perceive small quantities rapidly and accurately \cite{kaufman1949discrimination, revkin2008does}. However, the fine-grained learning dynamics differed qualitatively between embodied and vision-only architectures.

Embodied models exhibited a strictly sequential acquisition pattern: the model first achieved high accuracy on number 1, then progressively extended competence to 2, 3, 4, and so on through the numerical range. Critically, numbers that had not yet been ``learned'' showed near-zero accuracy. The model either counted correctly or failed entirely, with minimal partial knowledge. This all-or-none pattern mirrors the developmental trajectory observed in children, who progress through distinct stages as ``one-knowers,'' ``two-knowers,'' ``three-knowers,'' and eventually ``cardinal principle knowers'' who understand that counting applies to all numbers \cite{wynn1990children, le2007one}. The pretrained embodied model successfully acquired numbers 1--9 but showed persistent difficulty with 10, suggesting that the numerical range boundary poses particular challenges even for well-performing models.

Vision-only models displayed a qualitatively different pattern resembling the Approximate Number System (ANS), the evolutionarily ancient capacity for imprecise numerical estimation shared by humans, primates, and many other species \cite{feigenson2004core, dehaene2011number}. Rather than sequential acquisition, these models showed partial accuracy across the entire numerical range from early in training, with approximate discrimination of all quantities including 9 and 10, albeit with substantial errors. This pattern suggests that vision-only models rely on continuous magnitude estimation rather than discrete counting procedures, capturing the ``more versus less'' distinctions characteristic of the ANS but failing to develop the precise, symbolic counting competence enabled by embodied grounding.

Together, these results establish that embodied learning provides fundamental advantages for numerical cognition through multiple complementary mechanisms. Motor grounding enables superior data efficiency, with embodied models achieving high accuracy from limited training examples while vision-only models remain dependent on large datasets. The learning dynamics differ qualitatively: embodied models exhibit a two-phase trajectory, initial visuomotor calibration followed by rapid numerical learning, that parallels developmental observations in children. At the representational level, embodied learning produces linearly organized number representations resembling the mental number line, focused visual attention on task-relevant objects, and sequential number acquisition mirroring the subset-knower to cardinal-principle-knower progression documented in developmental psychology. These convergent findings suggest that sensorimotor grounding functions not merely as an additional information channel, but as a structural prior that shapes learning dynamics, representational geometry, and developmental trajectory in ways that align artificial numerical learning with biological cognition.

\begin{figure*}[htbp]
\centering
\includegraphics[width=\textwidth]{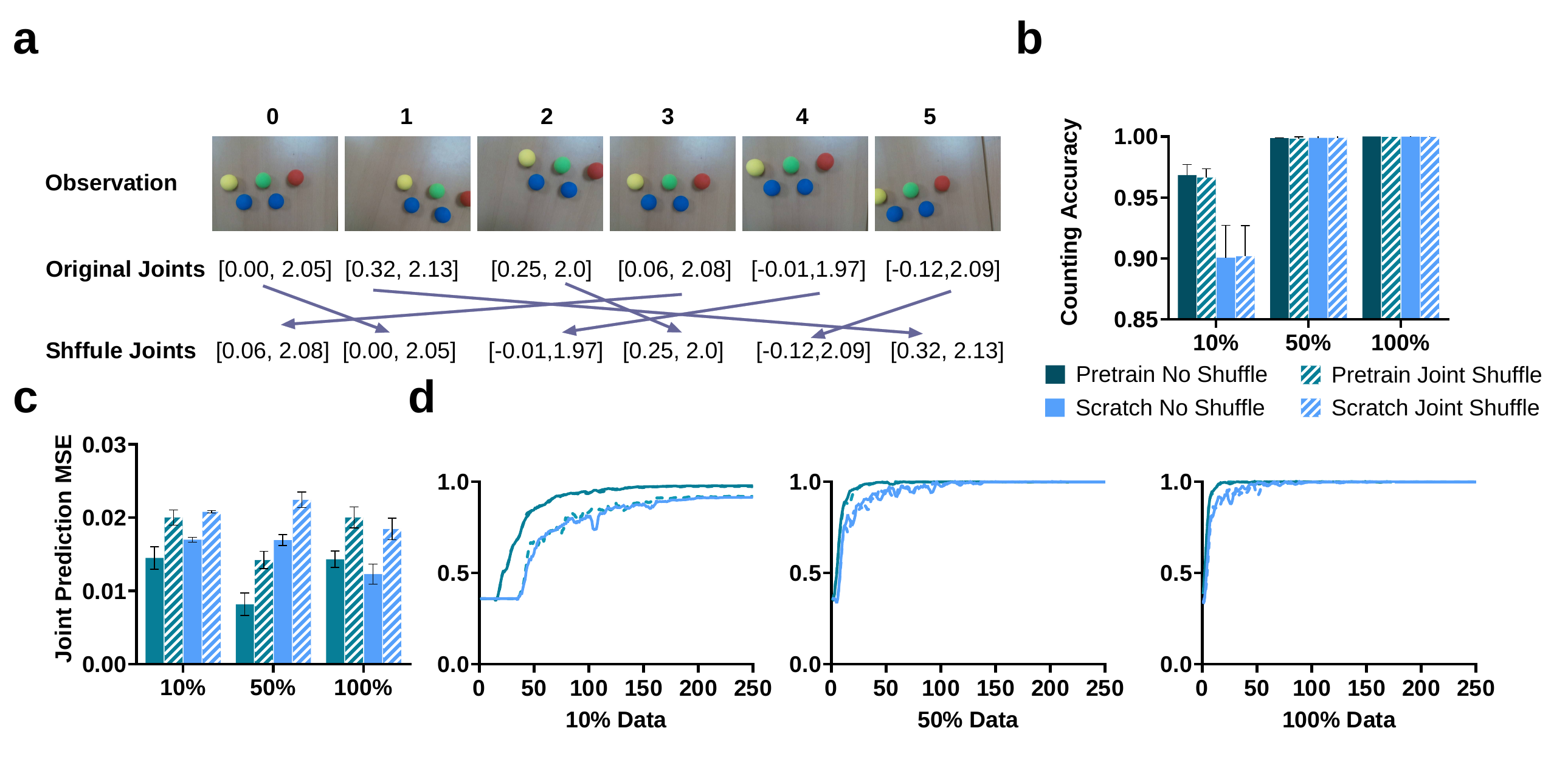}
\caption{\textbf{Motor signals function as structural priors rather than information sources.}
\textbf{a,} Schematic of the shuffle experiment. Within each training batch, joint position labels are randomly permuted across samples while visual inputs and count labels remain unchanged. This manipulation preserves motor prediction as an auxiliary training objective but eliminates veridical visuomotor correspondence.
\textbf{b,} Counting accuracy for intact (solid bars) versus shuffled (striped bars) visuomotor correspondence across data fractions. Shuffling has negligible effect on counting performance: models with disrupted correspondence achieve accuracy comparable to intact models across all conditions.
\textbf{c,} Motor prediction MSE for intact versus shuffled conditions. Unlike counting accuracy, motor prediction error increases substantially when visuomotor correspondence is disrupted, confirming that the shuffle manipulation successfully eliminated meaningful visual-motor relationships.
\textbf{d,} Learning trajectories for intact (solid lines) versus shuffled (dashed lines) conditions across data fractions (10\%, 50\%, 100\%). The characteristic two-phase learning profile, an initial latent period followed by rapid improvement, is preserved regardless of visuomotor correspondence. Together, panels b--d demonstrate a double dissociation: shuffling impairs motor prediction while leaving counting accuracy and learning dynamics intact, indicating that motor signals function as structural priors rather than information sources.}
\label{fig:shuffle_experiment}
\end{figure*}

\subsection{Motor signals function as structural priors rather than information sources}

The results above demonstrate that embodied learning substantially improves counting performance, but they do not yet explain why. Prior work has shown that finger representations can facilitate number learning in humanoid robots~\cite{Pecyna2020}. Notably, even random or unstructured motor signals have been reported to confer benefits over vision-only baselines~\cite{dinuovo2019developing}. This observation raises the question we posed earlier: does embodiment aid counting by providing task-relevant sensorimotor information, or by imposing structural constraints that regularize representation learning?

To dissociate these hypotheses, we conducted a shuffle experiment that selectively disrupted visuomotor correspondence while preserving motor supervision as an auxiliary learning objective. Within each training batch, joint position labels were randomly permuted across samples, leaving visual inputs and count labels unchanged (Figure~\ref{fig:shuffle_experiment}a). This manipulation preserves the presence of motor prediction as a training signal but eliminates any meaningful correspondence between visual observations and motor states. We compared embodied models trained with intact visuomotor alignment (E3 without pretraining, E9 with pretraining) to those trained with shuffled motor labels (E6 without pretraining, E12 with pretraining) across all data regimes (Table~\ref{tab:val_best}).

Shuffling motor labels had negligible effects on counting accuracy but consistently impaired motor prediction performance, revealing a clear double dissociation (Figure~\ref{fig:shuffle_experiment}b,c; Table~\ref{tab:val_best}). For counting accuracy, models with shuffled motor labels performed comparably to their intact counterparts across all conditions. At 10\% data, E6 (shuffled, non-pretrained) achieved 90.2\% ($\pm$2.5\%) accuracy, nearly identical to E3 (intact, non-pretrained) at 90.1\% ($\pm$2.7\%). Similarly, E12 (shuffled, pretrained) reached 96.6\% ($\pm$0.7\%), comparable to E9 (intact, pretrained) at 96.8\% ($\pm$0.9\%). This pattern persisted at 50\% and 100\% data fractions, with both intact and shuffled models achieving near-ceiling performance.

In contrast, motor prediction error increased substantially when visuomotor correspondence was disrupted (Figure~\ref{fig:shuffle_experiment}c). At 10\% data, MSE rose from 1.70$\times 10^{-2}$ (E3) to 2.08$\times 10^{-2}$ (E6) for non-pretrained models and from 1.46$\times 10^{-2}$ (E9) to 2.02$\times 10^{-2}$ (E12) for pretrained models. Similar increases were observed at 50\% and 100\% data fractions. Notably, pretrained models with intact correspondence showed the lowest MSE values, particularly at higher data fractions (0.69$\times 10^{-2}$ at 50\% data, 0.74$\times 10^{-2}$ at 100\% data for E9), indicating that these models successfully learned veridical visuomotor mappings. The shuffled models, by design, could not learn such mappings, resulting in persistently elevated prediction error.

Analysis of learning trajectories revealed that the temporal dynamics of counting acquisition were also preserved under shuffled conditions (Figure~\ref{fig:shuffle_experiment}d). Across all data fractions, models with shuffled motor labels exhibited the same characteristic two-phase learning profile observed in intact models: an initial latent period of approximately 20--50 epochs followed by rapid improvement toward ceiling accuracy. The learning curves for intact and shuffled conditions were nearly superimposed, indicating that the specific content of motor signals does not influence the temporal structure of numerical learning.

Together, these results reveal a double dissociation: disrupting visuomotor correspondence impairs motor prediction but leaves counting performance and learning dynamics intact. Embodiment thus improves counting not by supplying task-relevant sensorimotor information, but by introducing an auxiliary objective that functions as a structural prior, regularizing representation learning. This finding extends earlier observations~\cite{dinuovo2019developing} by clarifying the underlying mechanism: motor signals act as inductive biases that constrain learning dynamics, rather than as direct information sources for numerical cognition. This observation suggests that even minimal or symbolic forms of embodiment, without faithful sensorimotor coupling, may provide substantial learning benefits for abstract concept acquisition.

\begin{figure*}[htbp]
\centering
\includegraphics[width=\textwidth]{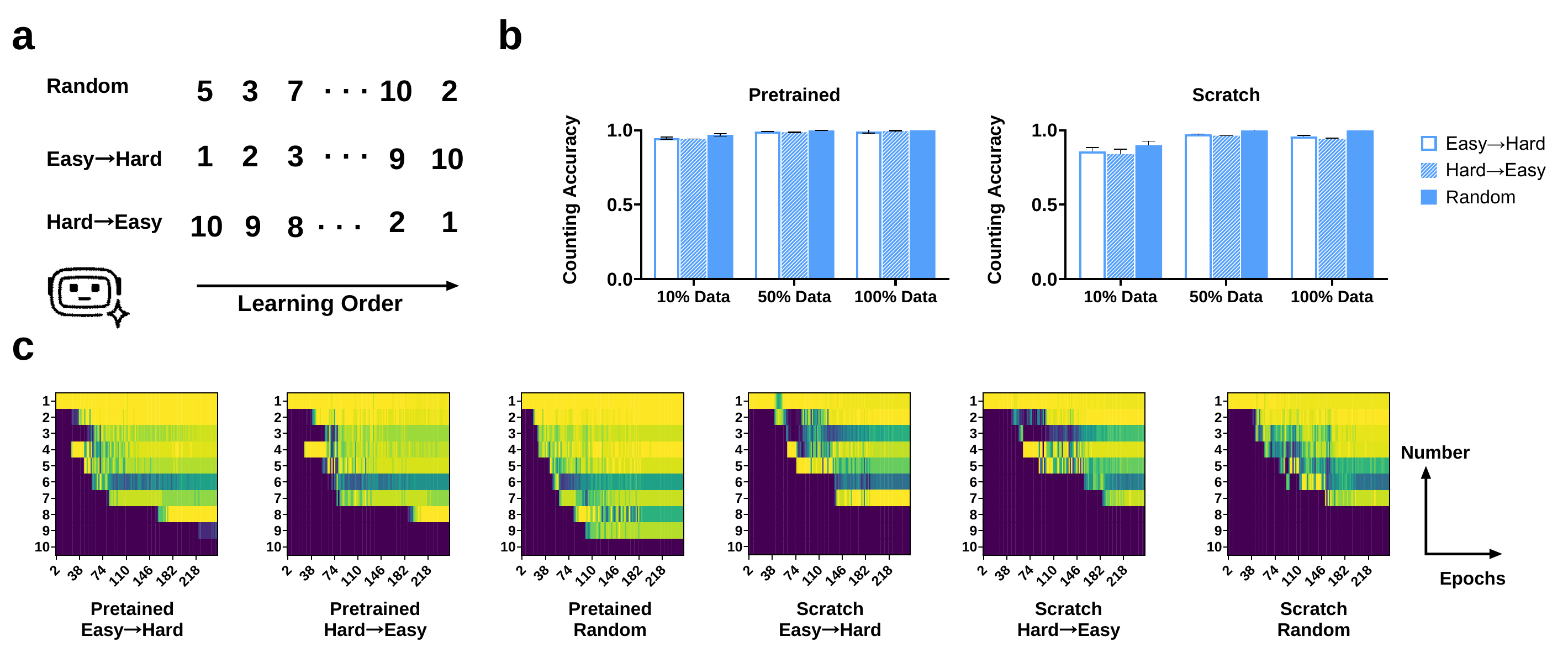}
\caption{\textbf{Curriculum learning reveals intrinsic developmental constraints in numerical acquisition.}
\textbf{a,} Schematic of three curriculum strategies. In easy-to-hard, training samples are ordered by numerosity from 1 to 10. In hard-to-easy, the order is reversed from 10 to 1. In random, samples are shuffled without systematic ordering.
\textbf{b,} Final counting accuracy across curriculum strategies and data fractions for pretrained (left) and non-pretrained (right) models. Random curriculum yields the best performance, followed by easy-to-hard, with hard-to-easy performing worst, particularly in low-data regimes. Error bars indicate standard deviation across 3 random seeds.
\textbf{c,} Heatmaps of per-number accuracy across training epochs for all curriculum and pretraining conditions (10\% data). Y-axis represents numerosity (1--10), X-axis represents training epochs, and color indicates accuracy (purple: low, yellow: high). Across all conditions, high accuracy (yellow) emerges first for small numbers (top rows) and progressively extends to larger numbers (bottom rows), regardless of curriculum strategy. This curriculum-invariant developmental pattern demonstrates that the small-to-large acquisition order reflects intrinsic computational constraints of the counting task rather than training order artifacts.}
\label{fig:curriculum}
\end{figure*}

\subsection{Learning trajectory mirrors human numerical development}

A hallmark of human numerical development is the systematic progression from small to large numbers. Children typically progress through distinct stages: they first become ``one-knowers,'' then ``two-knowers,'' and so on, before eventually grasping the cardinal principle that applies to all numbers~\cite{wynn1990children, le2007one}. This progression occurs regardless of the order in which numbers are presented in educational settings~\cite{sarnecka2009levels}. We examined whether our embodied model exhibits analogous developmental patterns and whether these patterns depend on the order of training examples.

We implemented three curriculum strategies that controlled the order of training samples within each epoch (Figure~\ref{fig:curriculum}a): easy-to-hard, in which samples were ordered by numerosity from 1 to 10; hard-to-easy, in which the order was reversed from 10 to 1; and random, in which samples were shuffled without systematic ordering. We trained embodied models under each curriculum strategy across all data fractions, comparing E1 (easy-to-hard, no pretraining), E2 (hard-to-easy, no pretraining), and E3 (random, no pretraining), as well as their pretrained counterparts E7, E8, and E9 (Table~\ref{tab:val_best}).

Curriculum strategy had modest but consistent effects on final counting performance (Figure~\ref{fig:curriculum}b). Random curriculum yielded the best results across most conditions, followed by easy-to-hard, with hard-to-easy performing worst. At 10\% data without pretraining, random curriculum (E3) achieved 90.1\% ($\pm$2.7\%) accuracy compared to 85.7\% ($\pm$2.7\%) for easy-to-hard (E1) and 84.0\% ($\pm$3.3\%) for hard-to-easy (E2). With pretraining, the pattern was similar: E9 (random) reached 96.8\% ($\pm$0.9\%), E7 (easy-to-hard) reached 94.7\% ($\pm$0.8\%), and E8 (hard-to-easy) reached 94.1\% ($\pm$0.1\%). These differences diminished at higher data fractions, where all curriculum strategies approached ceiling performance. The advantage of random ordering likely reflects better regularization through uniform exposure to all numerical categories throughout training, preventing representations from becoming overly specialized to particular numerical ranges.

The poor performance of hard-to-easy curriculum is particularly informative. When the model encounters only large numbers early in training, it cannot build stable representations because large-number counting depends on competencies that are more naturally acquired through small-number experience. This finding resonates with principles from incremental learning: complex skills are often best acquired by first mastering simpler component skills~\cite{elman1993learning, bengio2009curriculum}.

Despite these differences in final performance, analysis of per-number learning trajectories revealed a curriculum-invariant developmental pattern (Figure~\ref{fig:curriculum}c). Across all curriculum strategies and pretraining conditions, small numbers (1--4) were consistently acquired before large numbers. The heatmaps show that high accuracy (yellow) emerges first in the top rows (small numbers) and progressively extends to bottom rows (larger numbers) as training proceeds. This pattern holds even under hard-to-easy curriculum, where the model is explicitly presented with large numbers first: despite early exposure to 10, 9, 8..., the model still acquires competence for 1, 2, 3... before achieving accuracy on larger numerosities. Per-number learning trajectories for all embodied models across experimental conditions are shown in Supplementary Figure~\ref{fig: apeedix4}.

This curriculum-invariant developmental trajectory suggests that the difficulty gradient from small to large numbers reflects intrinsic computational constraints of the counting task rather than an artifact of training order. Several factors may contribute to this pattern. Small numbers require fewer sequential steps in the counting procedure, involve less complex visual scenes, and are more frequently represented in our Zipf-distributed dataset. The embodied counting task may also impose architectural constraints: accurate counting of N objects requires reliable processing of all sequences of length 1 through N, making small-number competence a prerequisite for large-number success.

The parallel with human development is notable. Children show this same small-to-large progression even when exposed to numbers in varied orders~\cite{le2007one, sarnecka2009levels}, suggesting that similar computational constraints may shape numerical learning in both biological and artificial systems. The ``subset-knower'' stages documented in developmental psychology~\cite{wynn1992children, le2007one} may reflect not pedagogical choices but fundamental constraints on how counting procedures can be acquired, constraints that our embodied model recapitulates despite its very different substrate.

Together with the results from Section~\ref{subsection A} demonstrating that embodied models acquire numbers sequentially while vision-only models show approximate discrimination across all numbers, these findings indicate that embodied learning not only improves data efficiency but also produces learning dynamics that align with human cognitive development. The emergence of human-like developmental trajectories in an artificial system trained on naturalistic counting data suggests that these patterns may reflect general computational principles of numerical cognition rather than human-specific biological constraints.

\begin{figure*}
    \centering
    \includegraphics[width=\textwidth]{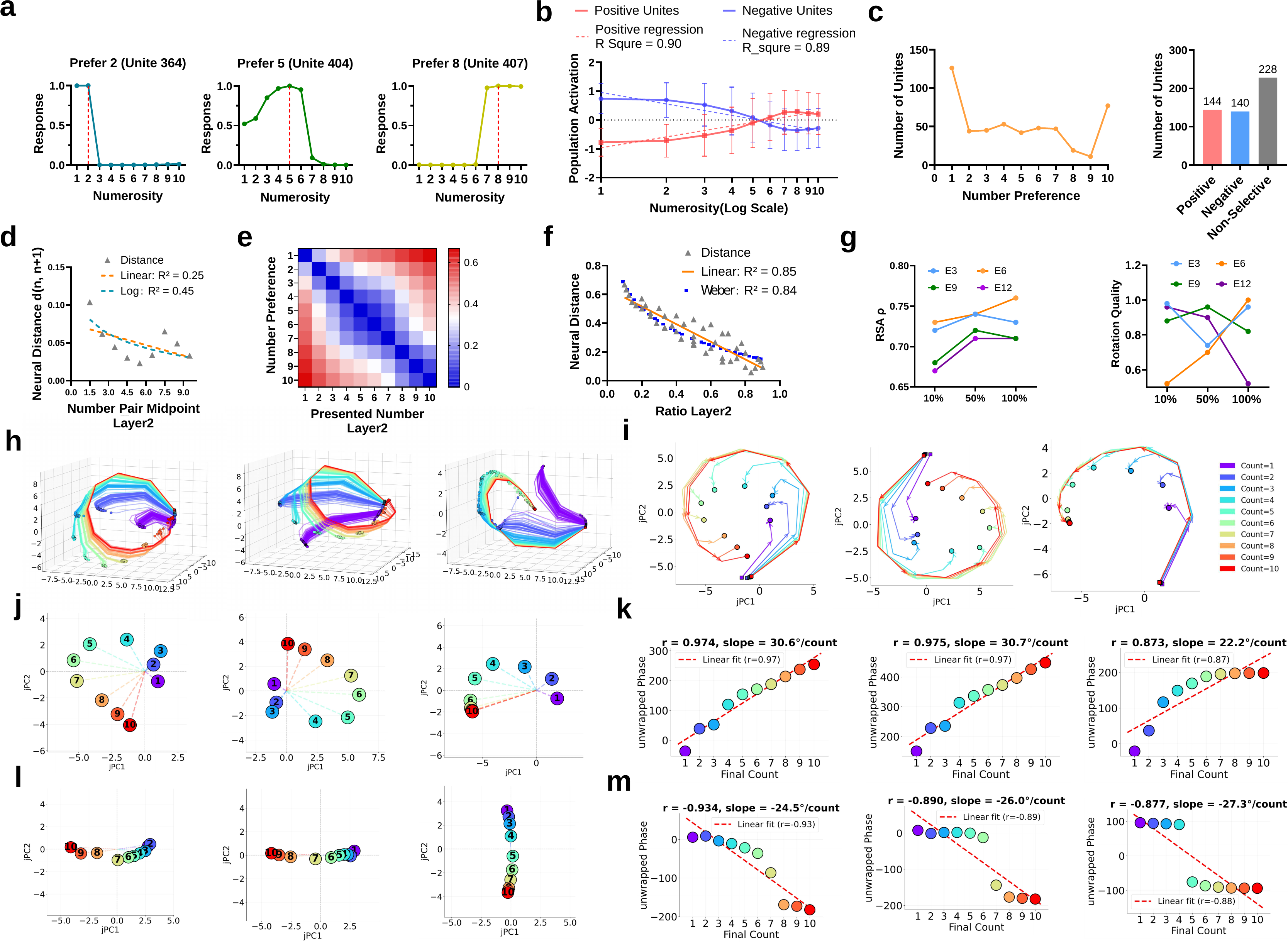}
   \caption{\textbf{Neural population dynamics reveal number-line organization and rotational counting mechanism.}
\textbf{(a)} Tuning curves of representative number-selective units preferring numerosities 2, 5, and 8.
\textbf{(b)} Population activity of positive (red) and negative (blue) numerosity detectors as a function of log(numerosity), showing logarithmic encoding ($R^2 = 0.90$ and $0.89$, respectively).
\textbf{(c)} Distribution of preferred numerosities across units (left) and proportion of positive, negative, and non-selective units (right).
\textbf{(d)} Neural distance between adjacent numbers decreases with magnitude, consistent with compressed number representations (logarithmic fit $R^2 = 0.45$).
\textbf{(e)} Representational dissimilarity matrix (RDM) showing graded similarity structure along the number line.
\textbf{(f)} Neural distance scales with numerical ratio (min/max), with both linear ($R^2 = 0.85$) and Weber-law ($R^2 = 0.84$) models providing good fits.
\textbf{(g)} RSA correlation (left) and rotation quality (right) remain stable across training conditions (E3, E6, E9, E12) and data fractions.
\textbf{(h)} PCA trajectories in Layer1 showing rotational dynamics at 100\%, 50\%, and 10\% training data.
\textbf{(i)} jPCA projections reveal clear rotational structure across data conditions.
\textbf{(j)} Terminal positions in jPCA space show orderly arrangement by numerical magnitude.
\textbf{(k)} Phase encoding: terminal jPCA phase correlates linearly with final count (100\%: $r = 0.974$, slope $= 30.6^\circ$/count; 50\%: $r = 0.975$; 10\%: $r = 0.873$), demonstrating that counting corresponds to traversing a single rotational cycle.}
\label{fig:neural_dynamics}
\end{figure*}

\subsection{Biologically plausible representations and dynamics emerge spontaneously}

Having established that embodied learning improves counting performance and produces human-like developmental trajectories, we next examined whether the trained model develops internal representations that parallel those observed in biological numerical cognition. We focused on three key properties documented in the neuroscience of number processing: number-selective tuning, spatial number line organization, and rotational population dynamics during sequential processing.

\textbf{Number-selective units emerge spontaneously.} We analyzed LSTM Layer 1 and 2 activations from the embodied model trained on 100\% data without pretraining (E3, random curriculum). Analysis of individual LSTM units revealed number-selective responses reminiscent of neurons in primate intraparietal cortex~\cite{nieder2006representation}. Figure~\ref{fig:neural_dynamics}a shows tuning curves of three representative units (Units 364, 404, and 407 in layer 1) that exhibit peaked responses to specific numerosities (2, 5, and 8, respectively), with graded falloff for neighboring numbers. Across the population, we identified two classes of numerosity detectors (Figure~\ref{fig:neural_dynamics}b): positive detectors whose activity increases monotonically with numerosity (144 units, 28.1\%), and negative detectors whose activity decreases with numerosity (140 units, 27.3\%). The remaining 228 units (44.5\%) showed no significant numerical selectivity (Figure~\ref{fig:neural_dynamics}c, right). Critically, both detector populations exhibited logarithmic scaling with numerosity ($R^2 = 0.90$ for positive detectors; $R^2 = 0.89$ for negative detectors), consistent with the logarithmic compression of numerical magnitude observed in human and primate neural recordings~\cite{dehaene2005three,nieder2025calculating}. The distribution of preferred numerosities showed boundary effects, with more units preferring extreme values (1 and 10) than intermediate numbers (Figure~\ref{fig:neural_dynamics}c, left), a pattern also observed in biological number neurons~\cite{nieder2006representation}.

\textbf{Number line organization in representational space.} A hallmark of numerical cognition in humans and primates is the mental number line---a spatial organization where numerically adjacent values occupy nearby positions in representational space~\cite{dehaene2005three,de2014representations}. We used representational similarity analysis (RSA) to test whether our model develops analogous structure. We computed representational dissimilarity matrices (RDMs) from LSTM Layer2 activations, measuring pairwise Euclidean distances between mean activation patterns for each numerosity (Figure~\ref{fig:neural_dynamics}e). The RDM exhibited a clear gradient structure: numerically similar quantities showed similar neural representations, while distant numbers were represented more distinctly.

Neural distances between adjacent numbers decreased systematically with numerical magnitude (Figure~\ref{fig:neural_dynamics}d), consistent with the logarithmic compression characteristic of Weber's law. Logarithmic fits ($R^2 = 0.45$) outperformed linear fits ($R^2 = 0.25$), indicating that representational precision decreases for larger numbers---a signature of approximate number representation in biological systems~\cite{nieder2025calculating,dehaene2005three}. Furthermore, neural distance scaled with numerical ratio (min/max) rather than absolute difference (Figure~\ref{fig:neural_dynamics}f), with both linear ($R^2 = 0.85$) and Weber-law ($R^2 = 0.84$) models providing excellent fits. This ratio-dependent discriminability is a defining feature of the approximate number system in humans and animals~\cite{feigenson2004core,gelman1986child}.

Across all training conditions, we observed significant positive correlations between neural distance and numerical distance (Spearman $\rho = 0.66$--$0.76$, all $p < 0.001$; Figure~\ref{fig:neural_dynamics}g, left; Table~\ref{tab:dynamics}), indicating that the LSTM spontaneously organizes number representations along a continuous dimension.This organization was remarkably robust: it emerged with only 10\% of training data ($\rho = 0.67$--$0.73$), was unaffected by shuffling visuomotor correspondence (L2, 10\% data: E3 vs E6, $\rho = 0.72$ vs $0.73$; E9 vs E12, $\rho = 0.68$ vs $0.67$), and persisted regardless of visual pretraining. The stability of number line organization across conditions suggests it reflects a fundamental computational solution for representing ordered magnitudes rather than an artifact of specific training configurations.

\textbf{Rotational dynamics during counting.} Recent work has revealed that neural population activity in motor cortex exhibits rotational dynamics during movement preparation and execution~\cite{churchland2012neural}. We applied jPCA analysis to examine whether similar dynamics emerge during sequential counting (Figure~\ref{fig:neural_dynamics}h--k). Following Churchland et al.~\cite{churchland2012neural}, we fit a linear dynamical system ($\dot{X} = MX$) to LSTM hidden states across counting sequences and decomposed the dynamics matrix $M$ into rotational (skew-symmetric) and non-rotational components.

Three-dimensional PCA trajectories of Layer1 hidden states revealed organized rotational structure during counting (Figure~\ref{fig:neural_dynamics}h). This structure persisted across data fractions (100\%, 50\%, 10\%), though trajectories became less organized with reduced training data. Strikingly, jPCA analysis revealed that rotational dynamics completely dominated the neural population activity: the skew-symmetric component explained 100\% of fitted dynamics variance across all conditions (Table~\ref{tab:dynamics}). The linear dynamical system provided good fits to the data, with $R^2$ ranging from 0.64 to 0.79 for non-pretrained models and 0.35 to 0.54 for pretrained models. Rotation quality scores ranged from 0.52 to 1.00, with most conditions showing strong rotational structure (quality $> 0.7$; Figure~\ref{fig:neural_dynamics}g, right).

Visualization of jPCA projections showed that trajectories for different numerosities followed similar rotational paths through neural state space but terminated at different phases (Figure~\ref{fig:neural_dynamics}i). Terminal positions in jPCA space were arranged in orderly fashion according to numerical magnitude (Figure~\ref{fig:neural_dynamics}j), with numbers progressing around the rotational plane. To quantify this relationship, we extracted the terminal phase angle for each numerosity and found a strong linear relationship between phase and count value (Figure~\ref{fig:neural_dynamics}k). At 100\% training data, the correlation was nearly perfect ($r = 0.974$), with a slope of $30.6^\circ$ per count. This relationship remained robust at 50\% data ($r = 0.975$, slope $= 30.7^\circ$/count) and was still significant at 10\% data ($r = 0.873$, slope $= 22.2^\circ$/count), though the reduced slope suggests incomplete rotational cycles with limited training.

The primary rotation frequency was approximately $\omega = 0.6$--$0.8$ rad/timestep, corresponding to a period of 8--10 counting steps---closely matching the numerical range of our task (1--10). This correspondence suggests that the LSTM implements counting as a single rotational cycle through neural state space, with the phase of rotation encoding counting progress. The model thus naturally implements the cardinal principle: the final rotational phase encodes the total count.

\textbf{Robustness to training manipulations.} Critically, both number line organization and rotational dynamics were preserved when visuomotor correspondence was shuffled (E6, E12; Table~\ref{tab:dynamics}). RSA correlations remained virtually identical between intact and shuffled conditions ($\Delta\rho < 0.02$), and rotational structure was fully maintained (100\% rotation variance in all shuffle conditions). This finding reinforces our central conclusion from Section B: motor signals function as structural priors that shape representation learning rather than as information sources. The specific content of motor signals is irrelevant; their presence as an auxiliary learning objective is sufficient to induce biologically plausible representational structure.

\begin{table*}[ht]
\centering
\caption{Neural dynamics and representational structure across training conditions for both LSTM layers. RSA $\rho$ = Spearman correlation between neural and numerical distance (all $p < 0.001$); $R^2$ = linear dynamics model fit quality; Quality = rotation quality score (0--1 scale). Rotational component explained 100\% of dynamics variance in all conditions. Random curriculum, Epoch 250.}
\label{tab:dynamics}
\resizebox{\textwidth}{!}{%
\begin{tabular}{ll ccc ccc ccc ccc}
\toprule
\textbf{Model} & \textbf{Layer}
& \multicolumn{3}{c}{\textbf{Conditions}}
& \multicolumn{3}{c}{\textbf{10\% Data}}
& \multicolumn{3}{c}{\textbf{50\% Data}}
& \multicolumn{3}{c}{\textbf{100\% Data}} \\
\cmidrule(r){3-5} \cmidrule(r){6-8} \cmidrule(r){9-11} \cmidrule(r){12-14}
& & \textbf{Pretrain} & \textbf{Shuffle} & \textbf{Curriculum}
& RSA $\rho$ & $R^2$ & Quality
& RSA $\rho$ & $R^2$ & Quality
& RSA $\rho$ & $R^2$ & Quality \\
\midrule
E3 & L1 & \texttimes & \texttimes & Random
  & 0.66 & 0.87 & 0.77  & 0.72 & 0.84 & 0.58  & 0.74 & 0.86 & 0.58 \\
E3 & L2 & \texttimes & \texttimes & Random
  & 0.72 & 0.77 & 0.55  & 0.74 & 0.79 & 0.68  & 0.74 & 0.64 & 0.84 \\[2pt]
E6 & L1 & \texttimes & \checkmark & Random
  & 0.67 & 0.87 & 0.73  & 0.72 & 0.84 & 0.56  & 0.73 & 0.86 & 0.52 \\
E6 & L2 & \texttimes & \checkmark & Random
  & 0.73 & 0.74 & 0.62  & 0.74 & 0.78 & 0.67  & 0.76 & 0.68 & 0.82 \\
\cmidrule(r){1-14}
E9 & L1 & \checkmark & \texttimes & Random
  & 0.70 & 0.83 & 0.67  & 0.72 & 0.82 & 0.63  & 0.70 & 0.85 & 0.64 \\
E9 & L2 & \checkmark & \texttimes & Random
  & 0.68 & 0.54 & 0.81  & 0.72 & 0.38 & 0.98  & 0.71 & 0.35 & 0.88 \\[2pt]
E12 & L1 & \checkmark & \checkmark & Random
  & 0.70 & 0.83 & 0.71  & 0.73 & 0.83 & 0.67  & 0.72 & 0.83 & 0.65 \\
E12 & L2 & \checkmark & \checkmark & Random
  & 0.67 & 0.48 & 0.94  & 0.71 & 0.41 & 0.96  & 0.71 & 0.38 & 0.58 \\
\bottomrule
\end{tabular}
}
\end{table*}

\textbf{Emergent computational principles.} Neither number line organization nor rotational dynamics were explicitly designed into the model architecture or training objective. The LSTM architecture imposes no preference for linear magnitude scaling or oscillatory state-space trajectories. Rather, these features emerged spontaneously from learning to count through embodied sensorimotor experience. This convergence between artificial and biological systems suggests that the computational demands of sequential enumeration, combined with embodied grounding, channel learning toward representational solutions that evolution has independently discovered for biological numerical cognition.

The emergence of rotational dynamics during counting is particularly noteworthy. Unlike the continuous reaching movements studied in Churchland et al.~\cite{churchland2012neural}, counting is a discrete sequential process. Yet the LSTM developed qualitatively similar rotational structure, with phase encoding task progress. This convergence suggests that rotational dynamics may represent a general computational motif for sequential processes---a neural ``clock'' that tracks progress through ordered sequences---rather than a mechanism specific to continuous motor control.

\section{Discussion}

This study investigated how embodied sensorimotor experience shapes numerical concept learning in artificial systems. We trained a neural network to perform sequential counting through robotic interaction and demonstrated three key advantages of embodiment: improved data efficiency, robust generalization, and spontaneous emergence of biologically plausible representations. These findings advance our understanding of embodied cognition and offer practical insights for designing efficient learning systems. The representational structures that emerged provide a computational framework for simulating human numerical cognition and testing cognitive theories. Our approach also demonstrates the value of applying neuroscientific methods such as jPCA and RSA to interpret artificial neural networks. This aligns with recent perspectives on using RNNs as formal surrogates for biological systems~\cite{durstewitz2023reconstructing}.

A central finding is that shuffling visuomotor correspondences produces nearly identical performance to intact embodiment. This suggests that embodiment functions as a structural prior rather than an information source. The motor prediction task imposes temporal coherence constraints on internal representations. This encourages the network to develop smooth, low-dimensional dynamics even when the motor-visual mapping is arbitrary. Recent work shows that RNNs perform computations through dynamic representational warping~\cite{pellegrino2025rnns}. Our results suggest a complementary view: motor tasks constrain networks to develop low-dimensional manifolds that facilitate numerical encoding.

Our findings build on developmental robotics work exploring embodiment and numerical cognition. Previous studies showed that proprioceptive feedback from robot fingers supports quicker formation of a uniform number line~\cite{dinuovo2019developing}, and that robots learning to count by pointing exhibit behaviour strikingly similar to children~\cite{Pecyna2020}. Electrophysiological evidence supports this link: observing finger-counting gestures interferes with arithmetic computation and activates the left angular gyrus~\cite{proverbio2019finger}. This tight neural coupling helps explain why parietal lesions can simultaneously cause finger agnosia and acalculia. Given this evidence, one might expect humanoid form to be essential for numerical cognition. Yet our non-humanoid manipulator developed similar representations without fingers. The critical factor appears to be the temporal structure of sensorimotor coordination, not anthropomorphic morphology.

The model spontaneously developed biologically plausible representations. Number-selective units showed logarithmic tuning, mirroring neurons in monkey parietal cortex~\cite{nieder2006representation}. These units organized into a mental number line following Weber's law, a signature documented across species~\cite{dehaene2005three}. Beyond static representations, jPCA revealed rotational dynamics encoding numerical magnitude. This resonates with recent findings that rotational planes in motor cortex vary systematically with reach direction~\cite{sabatini2024reach}. Our results suggest an analogous principle for abstract cognition: just as motor cortex uses rotation to generate movement sequences, our network uses rotation phase to encode count value. The model's developmental trajectory further paralleled children's progression from subset-knowers to cardinal principle knowledge~\cite{wynn1992children,le2007one}. Together, these convergences suggest that embodied learning captures fundamental computational challenges of numerical acquisition, independent of biological or artificial substrate.

This study has several limitations that point toward future directions. Our dataset follows a Zipf-like distribution with small numbers overrepresented. While this mirrors naturalistic statistics, it may confound developmental interpretations. Recent theoretical work suggests that key phenomena of numerical cognition, including the small-large discontinuity, may emerge from efficient representation under limited informational capacity~\cite{cheyette2020unified}. Testing whether our developmental patterns persist with balanced distributions would help disentangle learning dynamics from statistical regularities in the training data. We also tested only one robotic platform. Systematic comparison across morphologies with varying degrees of freedom and sensor modalities would clarify which aspects of embodiment are necessary versus sufficient. Our analysis focused on counting to ten; extending to larger numerosities, arithmetic, or symbolic manipulation would test the scalability of embodied learning. Finally, while we documented representational similarities to biological systems, direct comparisons between model activity and neural recordings would strengthen mechanistic claims. Such comparisons could leverage emerging frameworks for reconstructing computational dynamics from neural data~\cite{durstewitz2023reconstructing}.

Our findings open several directions for future research. The minimalist embodiment paradigm could extend to other abstract domains such as spatial reasoning or language acquisition~\cite{cohn2001qualitative,coventry2023spatial}. Our analytical framework combining jPCA and RSA could be applied to other architectures such as Transformers~\cite{vaswani2017attention}, potentially revealing shared computational motifs across systems. The developmental parallels we observed further suggest promising educational applications. Embodied learning strategies may enhance children's mathematics and science education~\cite{kosmas2019implementing,alibali1999function}, and robots trained with our approach could serve as interactive teaching tools. Notably, the activation of mirror neuron systems during action observation~\cite{rizzolatti2004mirror} suggests that observing a robot perform counting actions may engage neural processes similar to those involved in self-generated actions, potentially facilitating learning through observation. Finally, understanding how embodiment shapes numerical cognition may inform interventions for dyscalculia and other learning difficulties~\cite{butterworth2011dyscalculia}, leveraging sensorimotor grounding to support conceptual development.

\section{Method}
\label{sec:method}

\subsection{Model architecture}

We designed an embodied counting model that integrates visual perception with motor control through a recurrent architecture (Figure~\ref{fig:overview}D). The model comprises three components: a visual encoder, a motor encoder, a recurrent module, and dual output heads.

\paragraph{Visual encoder.}
We used AlexNet~\cite{krizhevsky2012imagenet} as the visual encoder. AlexNet is a convolutional neural network (CNN) inspired by the hierarchical organization of the primate visual cortex~\cite{hubel1968receptive,felleman1991distributed}. CNNs process images through alternating convolutional and pooling layers. A convolutional layer applies learnable filters to extract local features:
\begin{equation}
\mathbf{y}_{i,j} = \sigma\left(\sum_{m,n} \mathbf{W}_{m,n} \cdot \mathbf{x}_{i+m,j+n} + b\right)
\end{equation}
where $\mathbf{x}$ is the input, $\mathbf{W}$ is the filter, $b$ is the bias, and $\sigma$ is a nonlinear activation function (ReLU). Early layers detect simple features such as edges and textures, while deeper layers encode complex object representations. We extracted features from the final convolutional layer and applied global average pooling to obtain a 256-dimensional visual feature vector. We tested both randomly initialized and ImageNet-pretrained weights.

\paragraph{Motor encoder.}
A multilayer perceptron (MLP) encodes the robot's joint positions into a 256-dimensional motor feature vector. The motor encoder takes 2-DOF joint angles (joints 1 and 6, which capture the primary reaching motion) as input.

\paragraph{Recurrent module.}
The visual and motor features are concatenated to form a 512-dimensional input vector, which is processed by a two-layer Long Short-Term Memory (LSTM) network~\cite{hochreiter1997long} with 512 hidden units per layer and dropout ($p = 0.3$) between layers. Recurrent neural networks (RNNs) maintain a hidden state that integrates information across time, making them well-suited for sequential tasks. The LSTM addresses the vanishing gradient problem through gating mechanisms. At each timestep $t$, the LSTM updates its hidden state $\mathbf{h}_t$ and cell state $\mathbf{c}_t$ according to:
\begin{align}
\mathbf{f}_t &= \sigma(\mathbf{W}_f [\mathbf{x}_t; \mathbf{h}_{t-1}] + \mathbf{b}_f) \\
\mathbf{i}_t &= \sigma(\mathbf{W}_i [\mathbf{x}_t; \mathbf{h}_{t-1}] + \mathbf{b}_i) \\
\mathbf{o}_t &= \sigma(\mathbf{W}_o [\mathbf{x}_t; \mathbf{h}_{t-1}] + \mathbf{b}_o) \\
\tilde{\mathbf{c}}_t &= \tanh(\mathbf{W}_c [\mathbf{x}_t; \mathbf{h}_{t-1}] + \mathbf{b}_c) \\
\mathbf{c}_t &= \mathbf{f}_t \odot \mathbf{c}_{t-1} + \mathbf{i}_t \odot \tilde{\mathbf{c}}_t \\
\mathbf{h}_t &= \mathbf{o}_t \odot \tanh(\mathbf{c}_t)
\end{align}
where $\mathbf{x}_t$ is the 512-dimensional input, $\mathbf{f}_t$, $\mathbf{i}_t$, and $\mathbf{o}_t$ are the forget, input, and output gates, $\sigma$ is the sigmoid function, and $\odot$ denotes element-wise multiplication.

\paragraph{Output heads.}
Two linear heads operate on the LSTM hidden states. The counting head maps the final hidden state to a 10-way softmax distribution over numerosities 1--10. The motor prediction head outputs the 2-DOF joint positions at each timestep, providing an auxiliary training signal that encourages the network to track sensorimotor dynamics throughout the counting sequence.

\subsection{Training}

The model was trained with a combined loss function:
\begin{equation}
\mathcal{L} = \mathcal{L}_{\text{count}} + \lambda \mathcal{L}_{\text{motor}}
\end{equation}
where the counting loss is cross-entropy computed at the final timestep:
\begin{equation}
\mathcal{L}_{\text{count}} = -\sum_{k=1}^{10} y_k \log \hat{y}_k
\end{equation}
and the motor loss is mean squared error computed at every timestep:
\begin{equation}
\mathcal{L}_{\text{motor}} = \frac{1}{T} \sum_{t=1}^{T} \| \mathbf{m}_t - \hat{\mathbf{m}}_t \|^2
\end{equation}
Here $y_k$ is the ground-truth count label, $\hat{y}_k$ is the predicted probability, $\mathbf{m}_t$ and $\hat{\mathbf{m}}_t$ are the true and predicted joint positions at timestep $t$, and $T$ is the sequence length.

We used Adam optimizer with learning rate $10^{-4}$, weight decay $10^{-5}$, batch size 32, and trained for 250 epochs. We trained four embodied model variants by crossing two factors: visual initialization (random vs. ImageNet-pretrained) and motor loss ($\lambda = 0$ vs. $\lambda = 1$), yielding models E3, E6, E9, and E12 (Table~\ref{tab:val_best}).

To test whether embodiment benefits depend on veridical visuomotor correspondence, we created shuffled variants where motor sequences were randomly permuted across samples, breaking visual-motor correspondence while preserving the motor prediction task.

\subsection{Neural analysis}

\paragraph{Visual attention.}
We used Gradient-weighted Class Activation Mapping (Grad-CAM)~\cite{selvaraju2017grad} to visualize which image regions the model attends to during counting. Grad-CAM computes a weighted combination of feature maps:
\begin{equation}
L_{\text{Grad-CAM}} = \text{ReLU}\left(\sum_k \alpha_k A^k\right), \quad \alpha_k = \frac{1}{Z}\sum_i \sum_j \frac{\partial y^c}{\partial A^k_{ij}}
\end{equation}
where $A^k$ is the $k$-th feature map, $y^c$ is the score for class $c$, and $Z$ is the number of spatial locations.

\paragraph{Number-selective units.}
We identified units in the LSTM hidden layers that respond selectively to specific numerosities. For each unit $h$, we computed the mean activation for each count (1--10) to obtain a tuning curve, then calculated a selectivity index:
\begin{equation}
S_h = \frac{\sigma_h}{\mu_h + \epsilon}
\end{equation}
where $\sigma_h$ and $\mu_h$ are the standard deviation and mean of the tuning curve across counts. Units with selectivity above the 90th percentile were classified as number-selective.

\paragraph{Representational similarity analysis.}
We used RSA~\cite{kriegeskorte2008representational} to quantify whether the model's internal representations follow a mental number line organization. We computed pairwise Euclidean distances between mean LSTM activations for each numerosity pair, yielding a representational dissimilarity matrix (RDM). We then correlated this RDM with a theoretical numerical distance matrix $D_{ij} = |i - j|$ using Spearman's $\rho$.

\paragraph{Rotational dynamics.}
We applied jPCA~\cite{churchland2012neural} to characterize temporal dynamics in LSTM activity. After PCA preprocessing to $K$ dimensions, we fit a linear dynamical system $\dot{\mathbf{x}} = M\mathbf{x}$ and extracted the skew-symmetric component:
\begin{equation}
M_{\text{skew}} = \frac{1}{2}(M - M^\top)
\end{equation}
Eigendecomposition of $M_{\text{skew}}$ yields purely imaginary eigenvalues whose magnitudes indicate rotation frequency $\omega$. We projected trajectories onto the top jPCA plane and quantified rotation quality by measuring perpendicularity between position and velocity vectors.

\begin{acknowledgments}
This work was supported by the ERC and UKRI under the E-Talk project (Grant No. EP/Y029534/1). Computational resources were provided by the Computational Shared Facility (CSF3) at the University of Manchester. Code is available at \url{https://github.com/mrsgzg/ETALK_Counting.git}. Data is available at \url{https://osf.io/jk4u8/overview?view_only=95fcde69554045788995b8ab2fdabc0d}.
\end{acknowledgments}

\bibliographystyle{abbrv}
\bibliography{apssamp}

\clearpage

\section{Supplementary Material}
\begin{figure*}
    \centering
    \includegraphics[width=.5\linewidth]{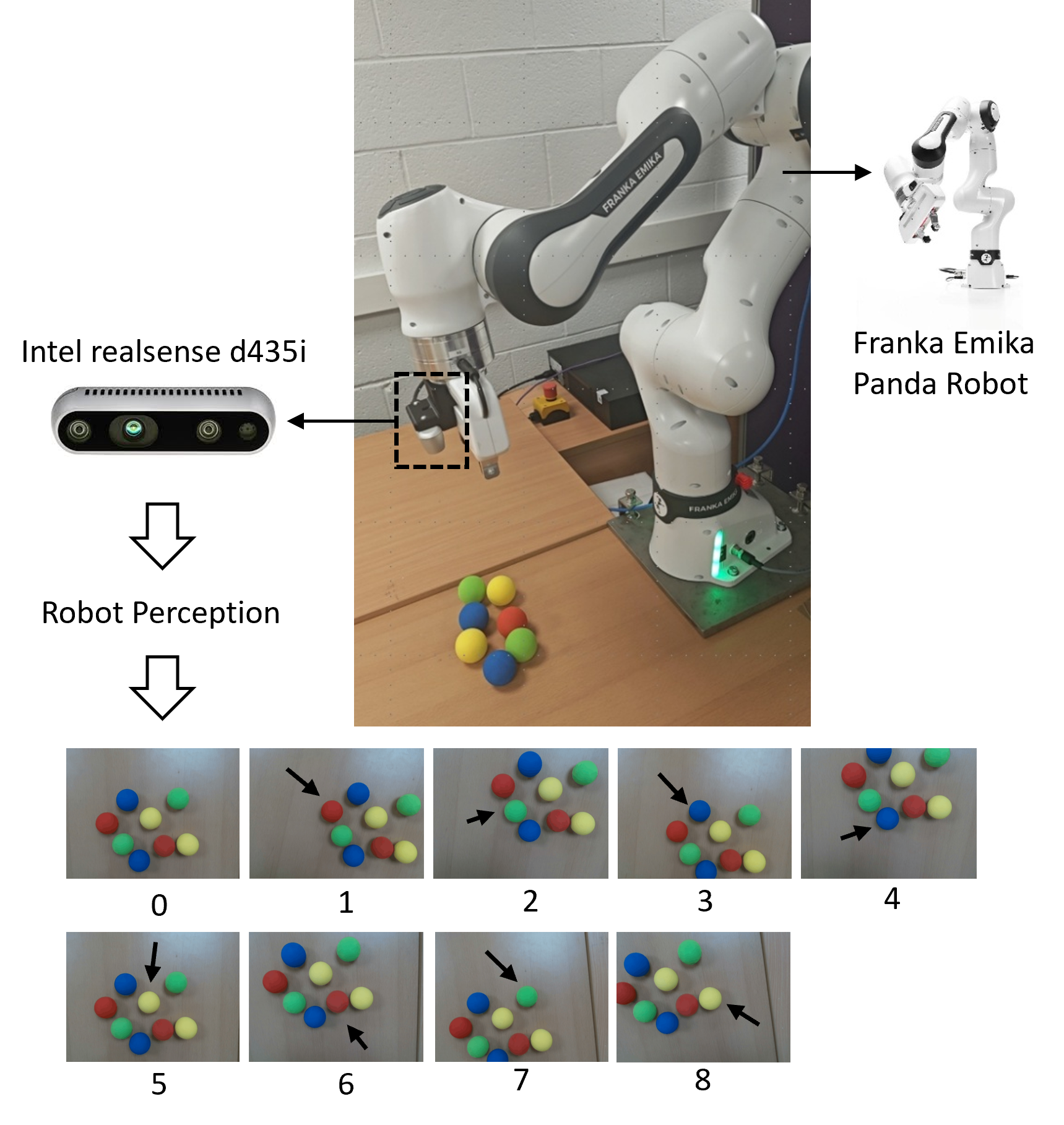}
    \caption{Overview of the experimental robotic platform. A Franka Emika Panda manipulator is equipped with an Intel RealSense D435i RGB-D camera for visual perception. The system performs object detection on colored balls (bottom-left inset) and controls the robot joints to sequentially reposition each object into the center of the camera’s field of view, enabling active perception and counting}
    \label{fig: apeedix1}
\end{figure*}

\subsection{Data collection}

We collected counting data using a Franka Emika Panda 7-DOF robotic manipulator equipped with a wrist-mounted Intel RealSense D435i RGB-D camera (Figure~\ref{fig: apeedix1}). The robot operated in a $90 \times 60$ cm workspace containing colored balls (diameter: 4 cm) arranged in various spatial configurations. For each counting trial, the robot sequentially visited each ball using predefined waypoints, and at each step we recorded an egocentric RGB image ($640 \times 480$ pixels) and the 7-DOF joint positions.

\begin{figure*}[h]
\centering
\includegraphics[width=0.5\linewidth]{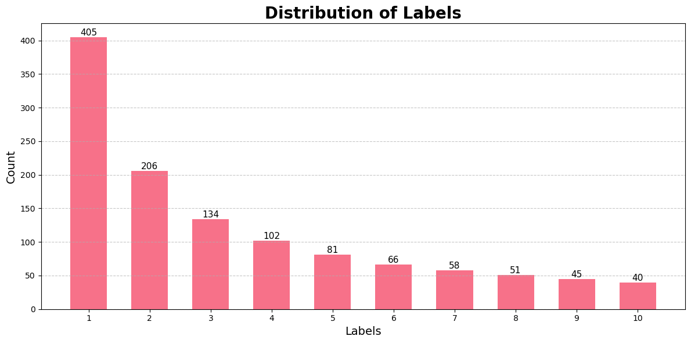}
\caption{\textbf{Dataset distribution.} Number of counting sequences for each numerosity (1--10), following a Zipf-like distribution.}
\label{fig:distribution}
\end{figure*}

The dataset comprises 1,188 counting sequences spanning numerosities 1 to 10. Each sequence contains $N$ image-motor pairs for a count of $N$.The distribution follows a Zipf-like pattern (Figure~\ref{fig:distribution}): 405, 206, 134, 102, 81, 66, 58, 51, 45, and 40 sequences for counts 1--10, respectively. Ball positions were randomized across trials to ensure spatial variability. We split the data into training and validation sets, with a fixed validation set of 242 samples. To examine data efficiency, we created training subsets of 94 (10\%), 472 (50\%), and 946 (100\%) samples.

\subsection{Baseline models}

To isolate the contribution of embodiment, we trained vision-only baseline models that received visual input without motor information or motor prediction objectives.

\paragraph{Single-image classifier (V1, V3).}
This baseline classifies numerosity from the final image of each counting sequence. The model consists of the same AlexNet encoder followed by a two-layer MLP:
\begin{equation}
\hat{y} = \text{softmax}(\mathbf{W}_2 \cdot \text{ReLU}(\mathbf{W}_1 \mathbf{v}_T + \mathbf{b}_1) + \mathbf{b}_2)
\end{equation}
where $\mathbf{v}_T \in \mathbb{R}^{256}$ is the visual feature from the final frame. This baseline tests whether counting can be solved by recognizing the final scene without temporal integration. V1 uses random initialization; V3 uses ImageNet pretraining.

\paragraph{Sequence pooling classifier (V2, V4).}
This baseline processes the full image sequence but without recurrent temporal modeling. Visual features are extracted independently for each frame and aggregated via mean pooling:
\begin{equation}
\bar{\mathbf{v}} = \frac{1}{T} \sum_{t=1}^{T} \mathbf{v}_t
\end{equation}
The pooled feature $\bar{\mathbf{v}} \in \mathbb{R}^{256}$ is then passed to the same MLP classifier. This baseline tests whether temporal order matters for counting. V2 uses random initialization; V4 uses ImageNet pretraining.

\begin{figure*}
    \centering
    \includegraphics[width=\linewidth]{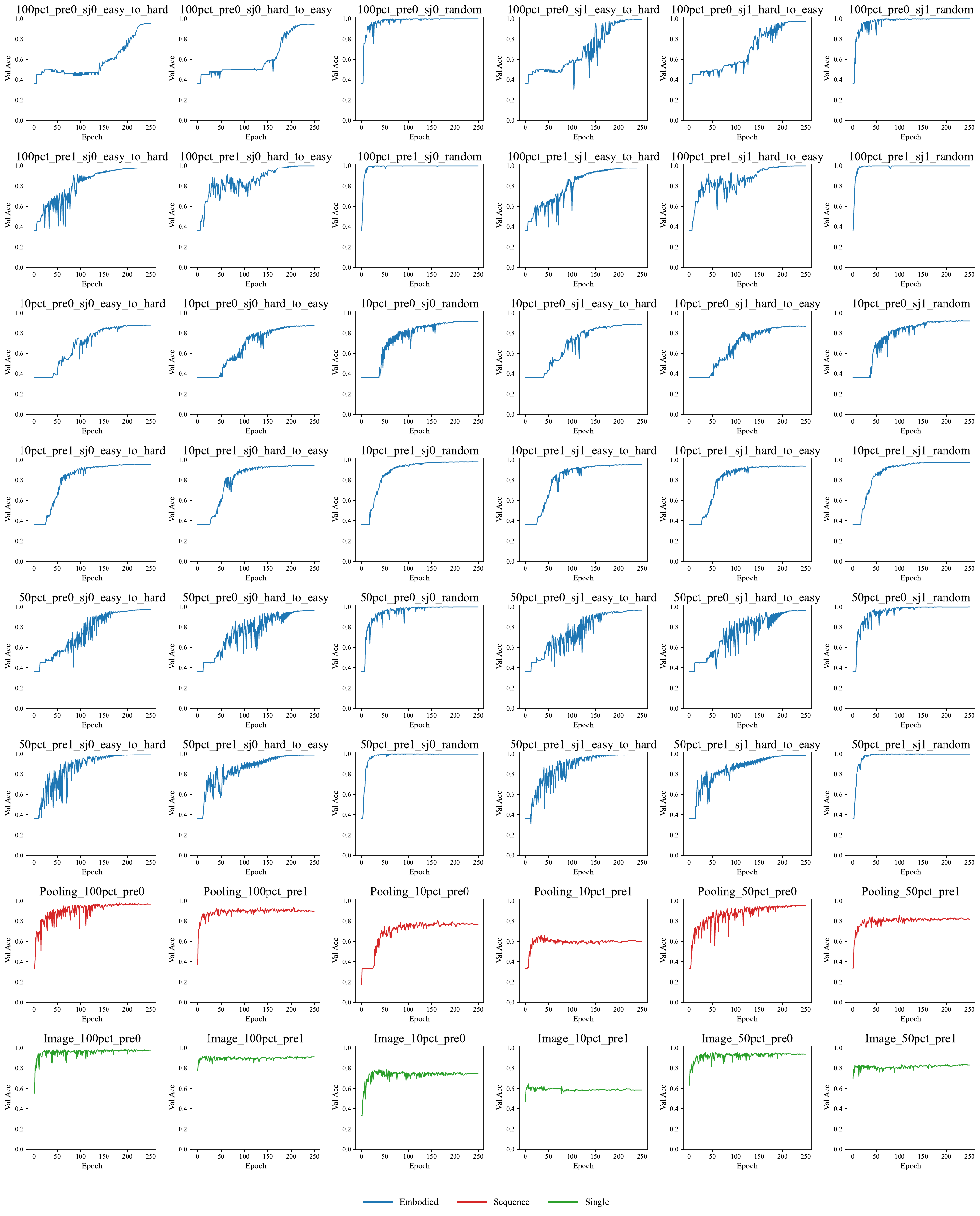}
    \caption{Learning trajectory of all training configurations, presented as validation-accuracy curves over epochs in a unified multi-panel layout (six panels per row). Panel titles indicate experimental settings, while color and legend denote model families.}
    \label{fig: apeedix3}
\end{figure*}

\begin{figure*}
    \centering
    \includegraphics[width=\linewidth]{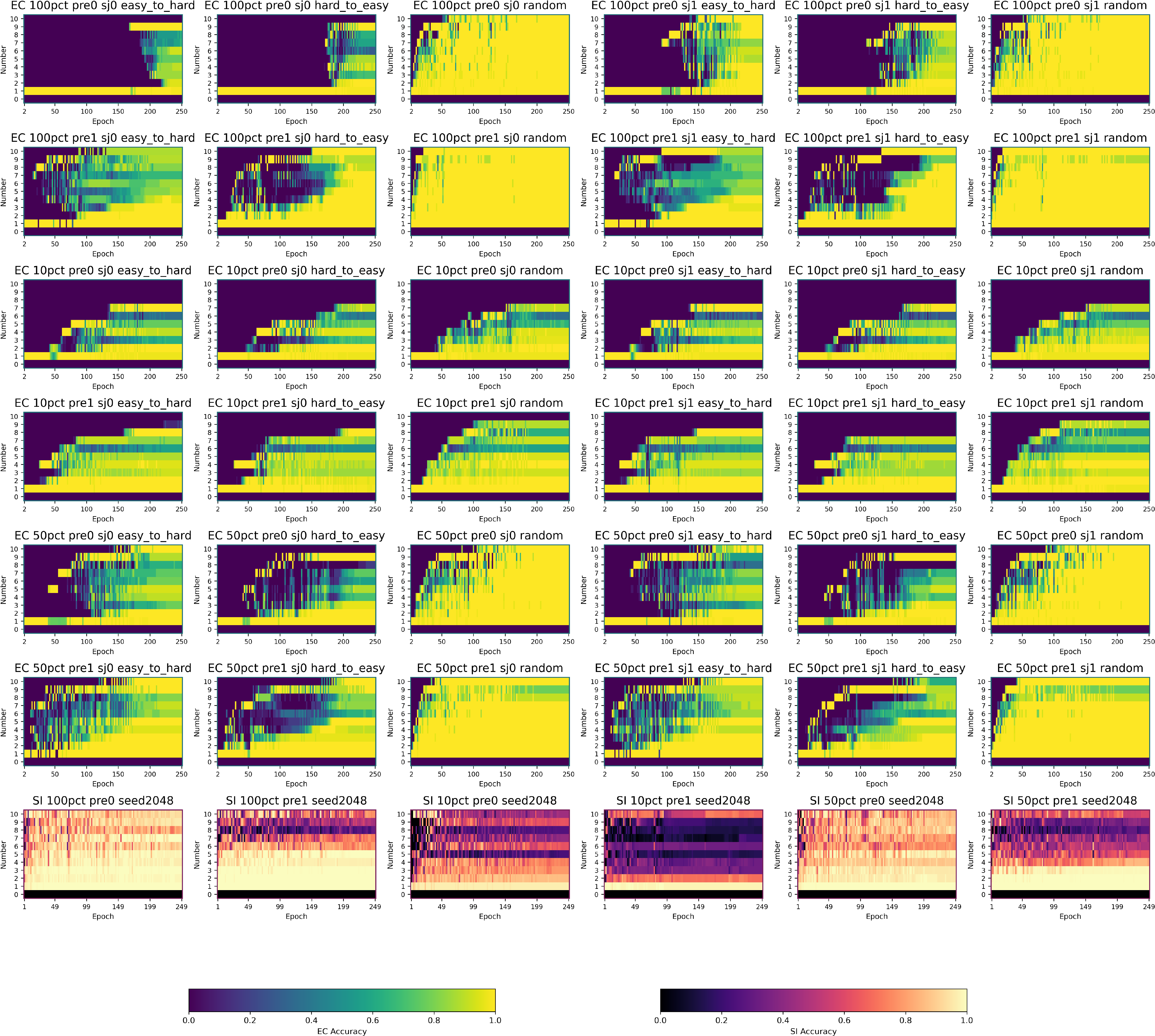}
    \caption{Heatmaps of per-number accuracy across training epochs for all experimental settings. Each panel corresponds to one result file, with epoch on the x-axis, number category (0–10) on the y-axis, and color indicating accuracy (0–1). Embodied Counting (EC) models (viridis colorbar, teal border) are labeled by training data proportion (10\%/50\%/100\%), pre-training condition (pre0/pre1), shuffle joint setting (sj0/sj1), and curriculum order (easy-to-hard, hard-to-easy, or random). Single Image (SI) models (magma colorbar, purple border) are labeled by training data proportion, pre-training condition, and random seed, and serve as a non-embodied baseline for comparison.}
    \label{fig: apeedix4}
\end{figure*}

\begin{figure*}[t]
\centering
\includegraphics[width=\linewidth]{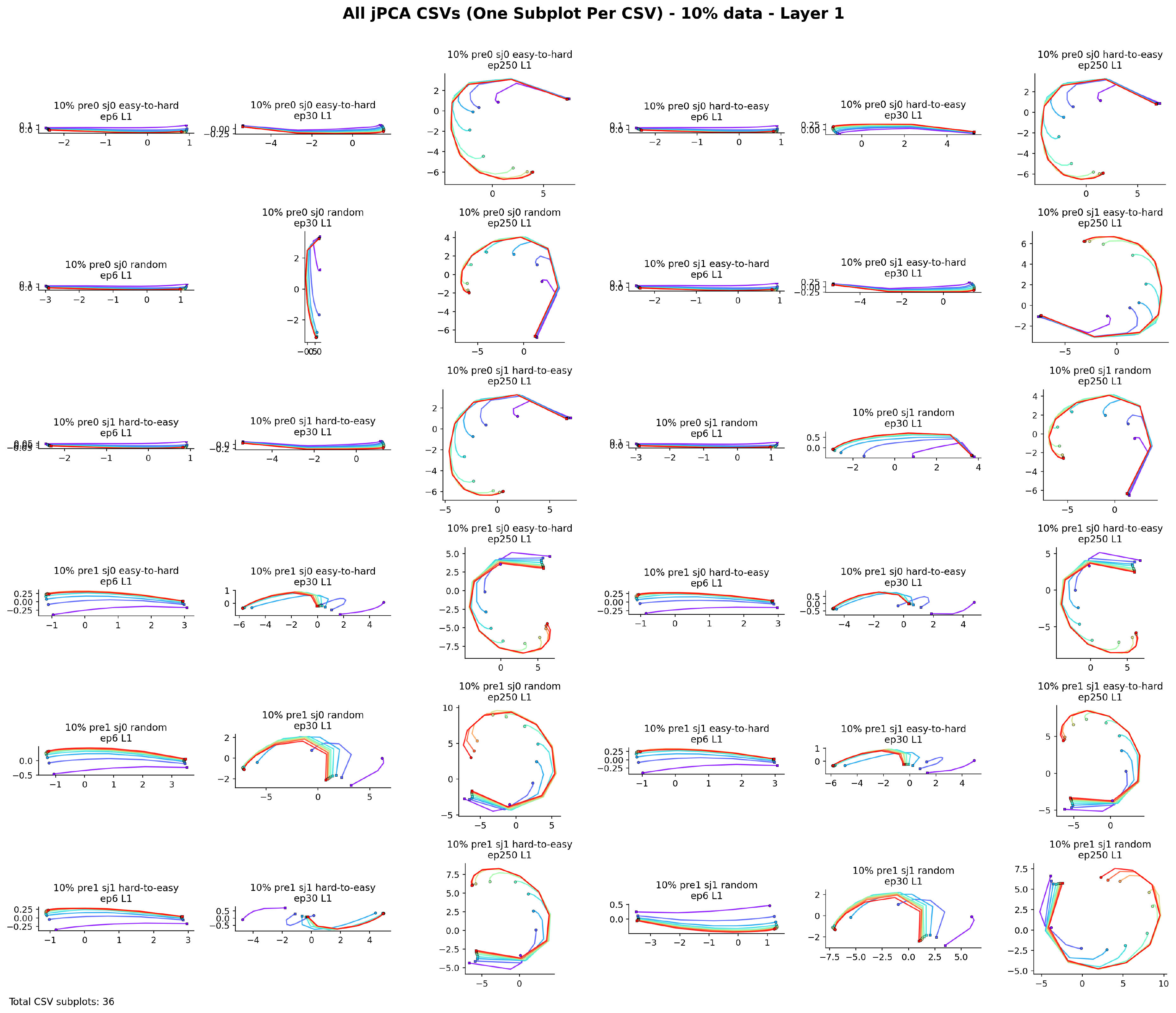}
\caption{Per-file jPCA trajectories for the 10\% data condition, layer1. Each subplot corresponds to one CSV result file (single model setting at a specific epoch), arranged in a 6-column grid. Within each subplot, colored curves denote count-specific trajectories in the jPC1-jPC2 plane. This panel enables fine-grained comparison of rotational structure under low-data training.}
\label{fig:jpca_10pct_layer1}
\end{figure*}

\begin{figure*}[t]
\centering
\includegraphics[width=\linewidth]{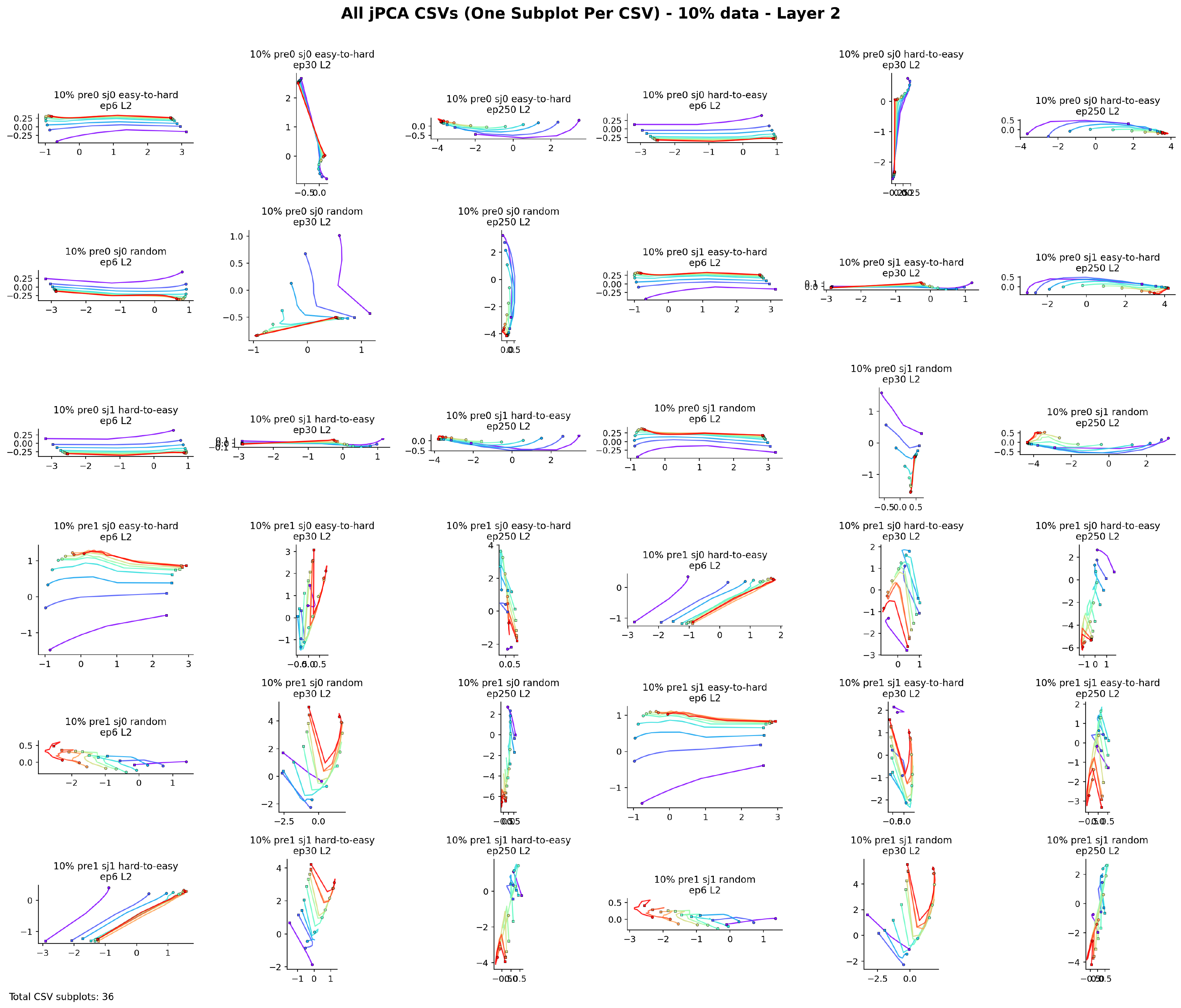}
\caption{Per-file jPCA trajectories for the 10\% data condition, layer2. Each subplot corresponds to one CSV result file, arranged in a 6-column grid. The panel highlights how rotational dynamics in layer2 vary across curriculum orders, pre-training conditions, shuffle-joint settings, and epochs when training data is limited.}
\label{fig:jpca_10pct_layer2}
\end{figure*}

\begin{figure*}[t]
\centering
\includegraphics[width=\linewidth]{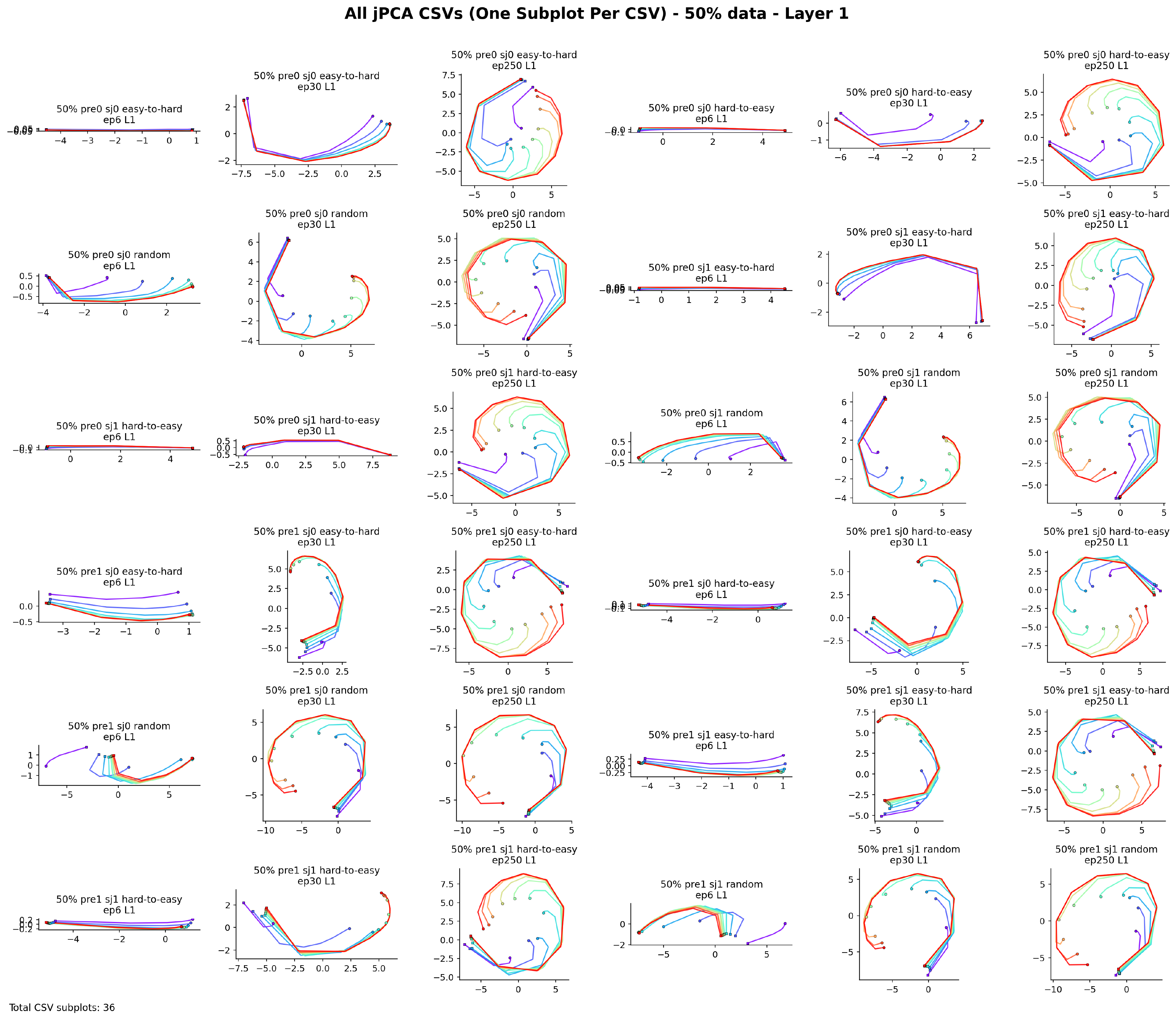}
\caption{Per-file jPCA trajectories for the 50\% data condition, layer1. Each subplot corresponds to one CSV result file, arranged in a 6-column grid. Compared with 10\% training, this panel allows inspection of increased trajectory stability and organization under intermediate data availability.}
\label{fig:jpca_50pct_layer1}
\end{figure*}

\begin{figure*}[t]
\centering
\includegraphics[width=\linewidth]{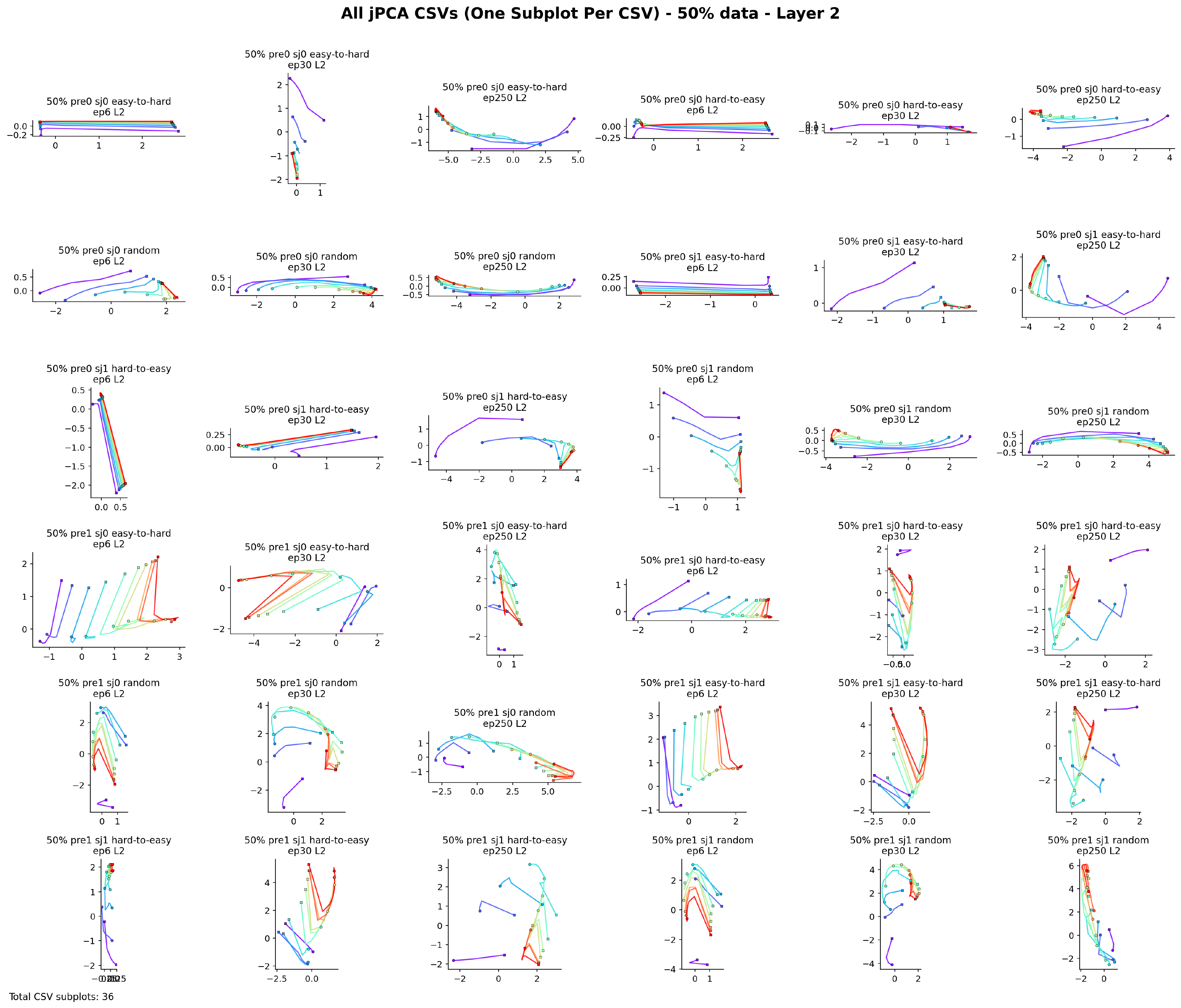}
\caption{Per-file jPCA trajectories for the 50\% data condition, layer2. Each subplot corresponds to one CSV result file, arranged in a 6-column grid. This panel provides a condition-wise view of layer2 rotational geometry at intermediate data scale across all curricula and training setups.}
\label{fig:jpca_50pct_layer2}
\end{figure*}

\begin{figure*}[t]
\centering
\includegraphics[width=\linewidth]{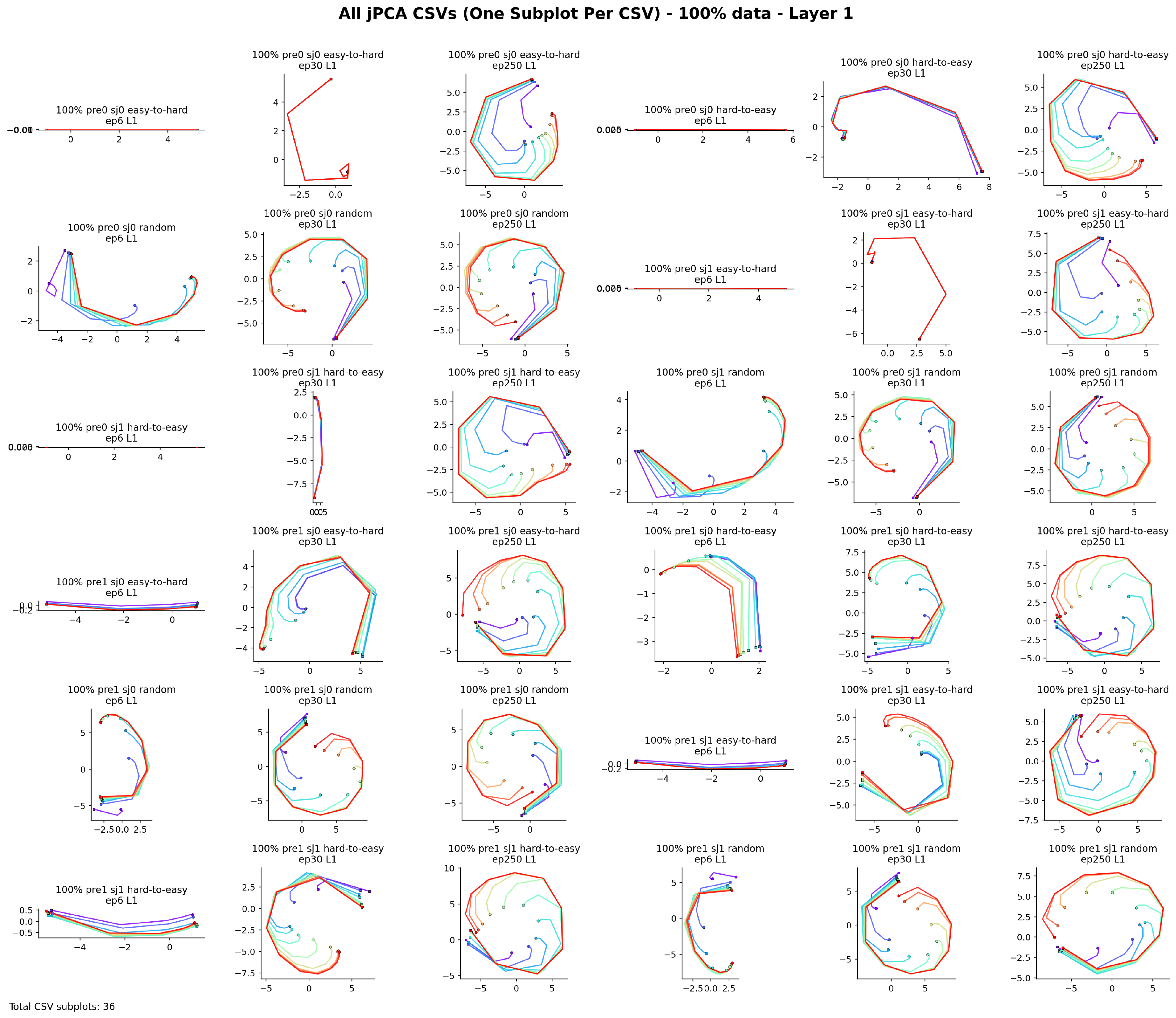}
\caption{Per-file jPCA trajectories for the 100\% data condition, layer1. Each subplot corresponds to one CSV result file, arranged in a 6-column grid. This panel captures full-data rotational dynamics and supports direct comparison with the 10\% and 50\% conditions.}
\label{fig:jpca_100pct_layer1}
\end{figure*}

\begin{figure*}[t]
\centering
\includegraphics[width=\linewidth]{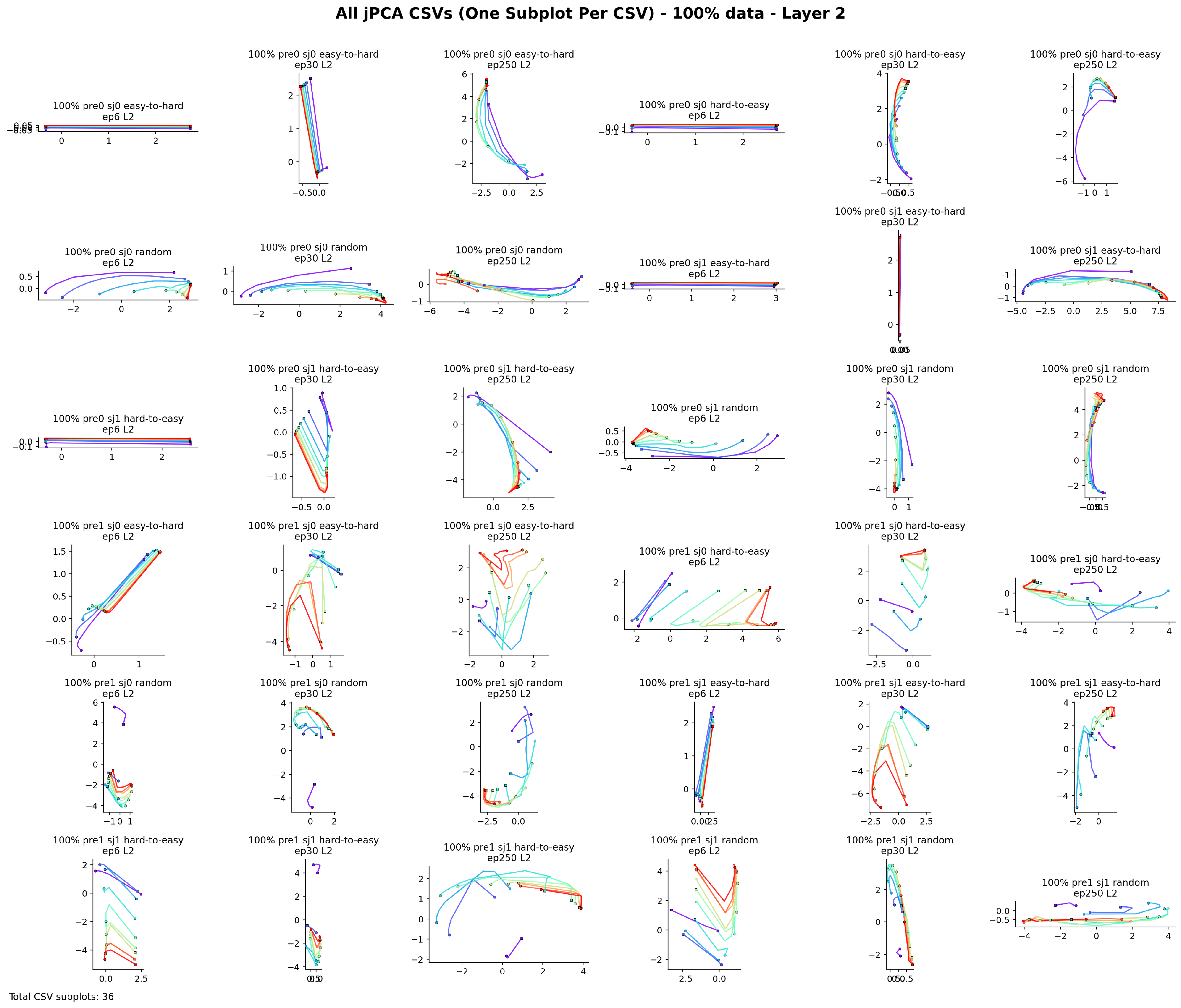}
\caption{Per-file jPCA trajectories for the 100\% data condition, layer2. Each subplot corresponds to one CSV result file, arranged in a 6-column grid. Together with the other five split panels and the global overlay figure, this panel completes the full comparison of jPCA rotational dynamics across data regimes, layers, and training configurations.}
\label{fig:jpca_100pct_layer2}
\end{figure*}

\nocite{*}

\end{document}